\documentclass[10pt,twocolumn,letterpaper]{article}

\usepackage{cvpr}
\usepackage{times}
\usepackage{epsfig}
\usepackage{graphicx}
\usepackage{amsmath}
\usepackage{amssymb}

\usepackage{caption} 
\usepackage{subcaption}
\usepackage{siunitx}
\usepackage{booktabs}
\usepackage{multirow} 
\usepackage{rotating}
\usepackage[sort,nocompress]{cite}
\usepackage[pagebackref=true,breaklinks=true,colorlinks,bookmarks=false]{hyperref}

\DeclareGraphicsExtensions{.pdf,.jpg,.png}
\graphicspath{{figs/}}

\newcommand{\figref}[1]{Figure~\ref{fig:#1}}
\newcommand{\tblref}[1]{Table~\ref{tbl:#1}}

\let\origfootnote\footnote
\renewcommand{\footnote}[1]{\kern.06em\origfootnote{#1}}
\newcommand{\punctfootnote}[1]{\kern-.06em\origfootnote{#1}}

\cvprfinalcopy 

\def\cvprPaperID{384} 

\begin{document}

\title{Learning a Dynamic Map of Visual Appearance}

\author{
    \centering
    \begin{minipage}{.9\linewidth}
      \centering
      \begin{minipage}{1.7in}
        \centering
        Tawfiq Salem\\
      \end{minipage}
      \begin{minipage}{1.7in}
        \centering
        Scott Workman\\
      \end{minipage}
      \begin{minipage}{1.7in}
        \centering
        Nathan Jacobs\\
      \end{minipage}
      \\[.2cm]
      \begin{minipage}{1.7in}
        \centering
        Miami University
      \end{minipage}
      \begin{minipage}{1.7in}
        \centering
        DZYNE Technologies
      \end{minipage}
      \begin{minipage}{1.7in}
        \centering
        University of Kentucky
      \end{minipage}
    \end{minipage}
}

\maketitle

\begin{abstract}
  The appearance of the world varies dramatically not only from place to place but also from hour to hour and month to month. Every day billions of images capture this complex relationship, many of which are associated with precise time and location metadata. We propose to use these images to construct a global-scale, dynamic map of visual appearance attributes. Such a map enables fine-grained understanding of the expected appearance at any geographic location and time. Our approach integrates dense overhead imagery with location and time metadata into a general framework capable of mapping a wide variety of visual attributes. A key feature of our approach is that it requires no manual data annotation. We demonstrate how this approach can support various applications, including image-driven mapping, image geolocalization, and metadata verification. 
\end{abstract}

\section{Introduction} \label{sec:intro}

\begin{figure}
    \centering    
    \includegraphics[width=\linewidth]{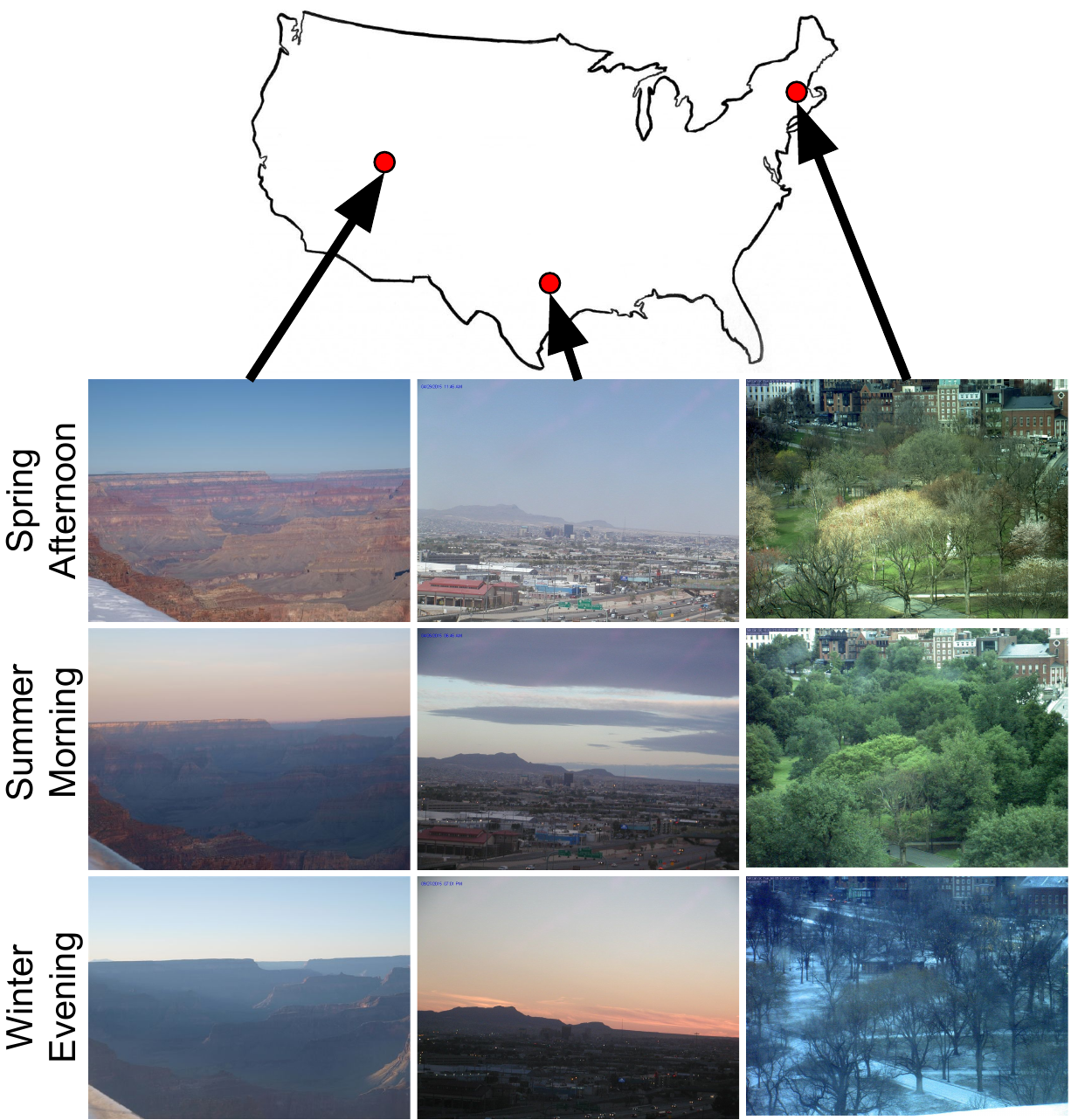}
    \caption{Visual appearance changes dramatically due to differences in location and time. Our work takes advantage of sparsely distributed ground-level image data, with associated location and time metadata, in conjunction with overhead imagery to construct dynamic maps of visual appearance attributes.}
    
    \label{fig:cartoon}
    
\end{figure}

Recent concern about ``fake news'' has lead to a significant interest in verifying that imagery is real and un-manipulated. Early work on this problem focused on low-level image statistics~\cite{farid2009image, bianchi2012image}, but this approach is unable to detect the falsification of image metadata. Matzen and Snavely~\cite{matzen2014scene} introduce an approach for finding anomalous timestamps, but their method is based on visual correspondences and requires overlapping imagery. Recent work has begun to look at this problem more thoroughly, with new datasets~\cite{guan2019mfc} and proposals for comprehensive systems~\cite{bharati2019beyond}. However, no previous work provides the dynamic map of visual attributes that is necessary for detecting time/location metadata falsification.

We propose to use visual attributes estimated from ground-level images, such as those shown in \figref{cartoon}, to learn a dynamic map of visual attributes. Beyond metadata verification, there are numerous applications for such a map, including geolocalizing images, providing contextual information for autonomous vehicles, and supporting further studies on the relationship between the visual environment and human health and happiness~\cite{seresinhe2015quantifying}.

Predicting visual attributes directly from location and time is difficult because of the complexity of the distribution. It would, for example, require memorizing the location of every road and building in the area of interest. To overcome this, our model combines overhead imagery with location and time using a multi-modal convolutional neural network. The result is a model capable of generating a worldwide, dynamic map of visual attributes that captures both local and global patterns.

We focus on two visual attributes: the scene category~\cite{zhou2017places}, such as whether the image views an attic or a zoo, and transient attributes~\cite{laffont2014transient}, which consist of time-varying properties such as sunny and foggy. We selected these because they are well known, easy to understand, and have very different spatiotemporal characteristics. The former is relatively stable over time, but can change rapidly with respect to location, especially in urban areas. The latter has regular, dramatic changes throughout the day and with respect to the season. 

Our approach has several useful properties: it does not require any manually annotated training data; it can model differences in visual attributes at large and small spatial scales; it captures spatiotemporal trends, but does not require overhead imagery at every time; and is extendable to a wide range of visual attributes. To evaluate our approach, we created a large dataset of paired ground-level and overhead images each with location and time metadata, which we call Cross-View Time (CVT). Using CVT, we demonstrate the effectiveness of our dynamic mapping approach on several tasks. In each case, our full model, which combines overhead imagery and metadata, is superior.
\section{Related Work}

Recent advances in computer vision have enabled the estimation of a wide variety of visual attributes, including scene category~\cite{zhou2017places}, weather conditions~\cite{laffont2014transient}, and demographics~\cite{gebru2017using}. As these techniques mature, many application areas have developed that require an understanding of the relationship between visual attributes, geographic location, and time. 

\begin{figure*}
    \centering
    \newcommand{\centered}[1]{\begin{tabular}{l} #1 \end{tabular}}
    \setlength\tabcolsep{1pt}
    \begin{subfigure}{0.216\textwidth}
        \centering
        \begin{tabular}{c}
            \includegraphics[width=1\linewidth]{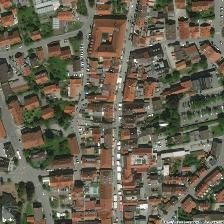}
        \end{tabular}
    \end{subfigure}
    \begin{subfigure}{0.744\textwidth}
      \begin{tabular}{cccc}
          \includegraphics[width=.249\linewidth]{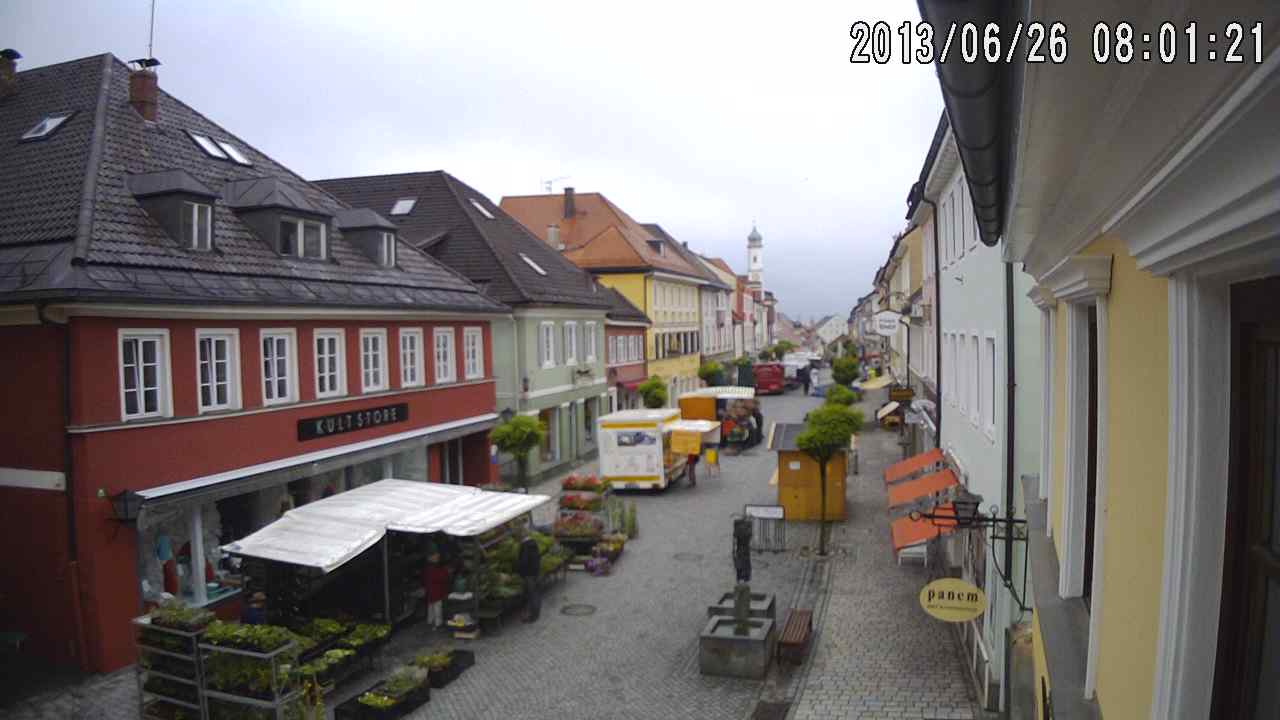} &
          \includegraphics[width=.249\linewidth]{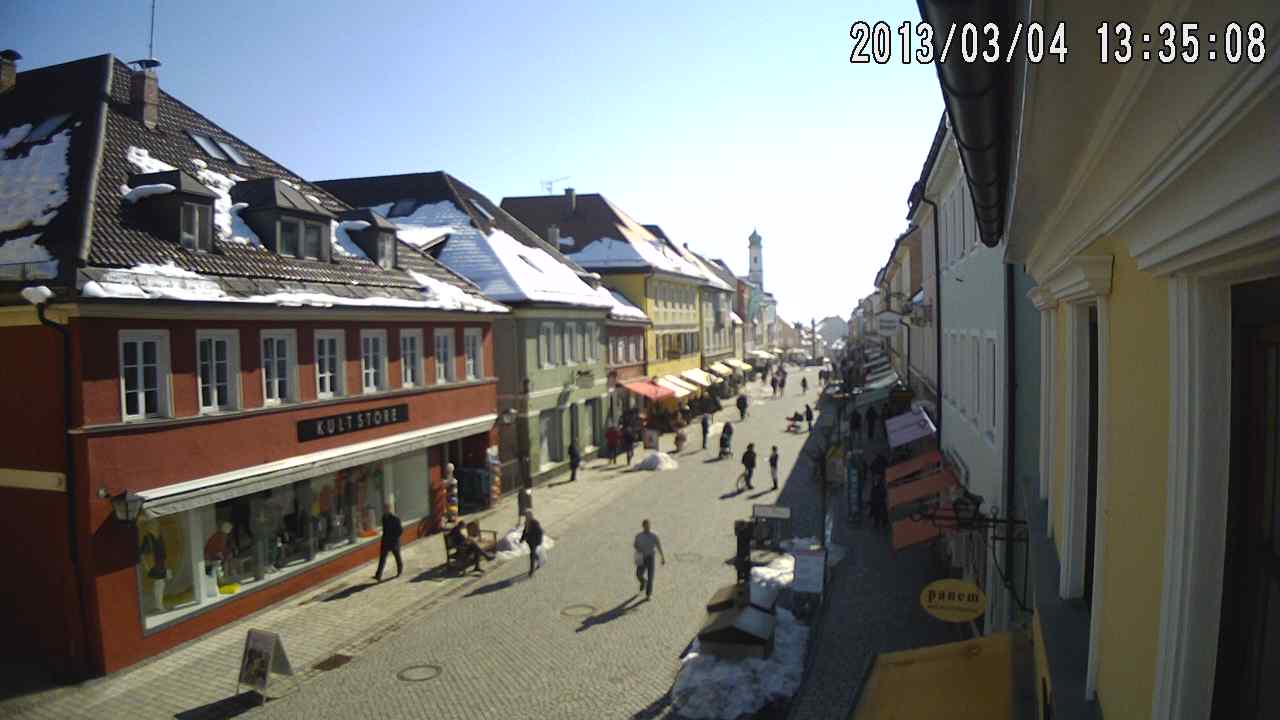} &
          \includegraphics[width=.249\linewidth]{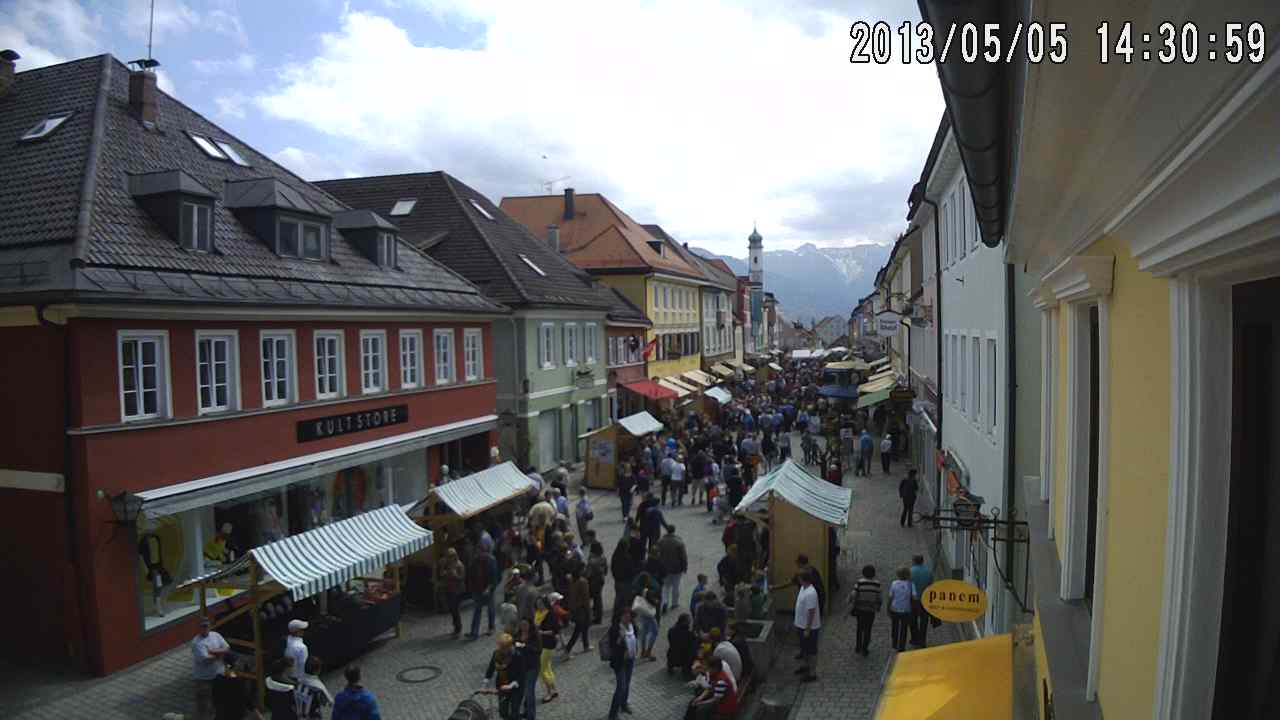} &
          \includegraphics[width=.249\linewidth]{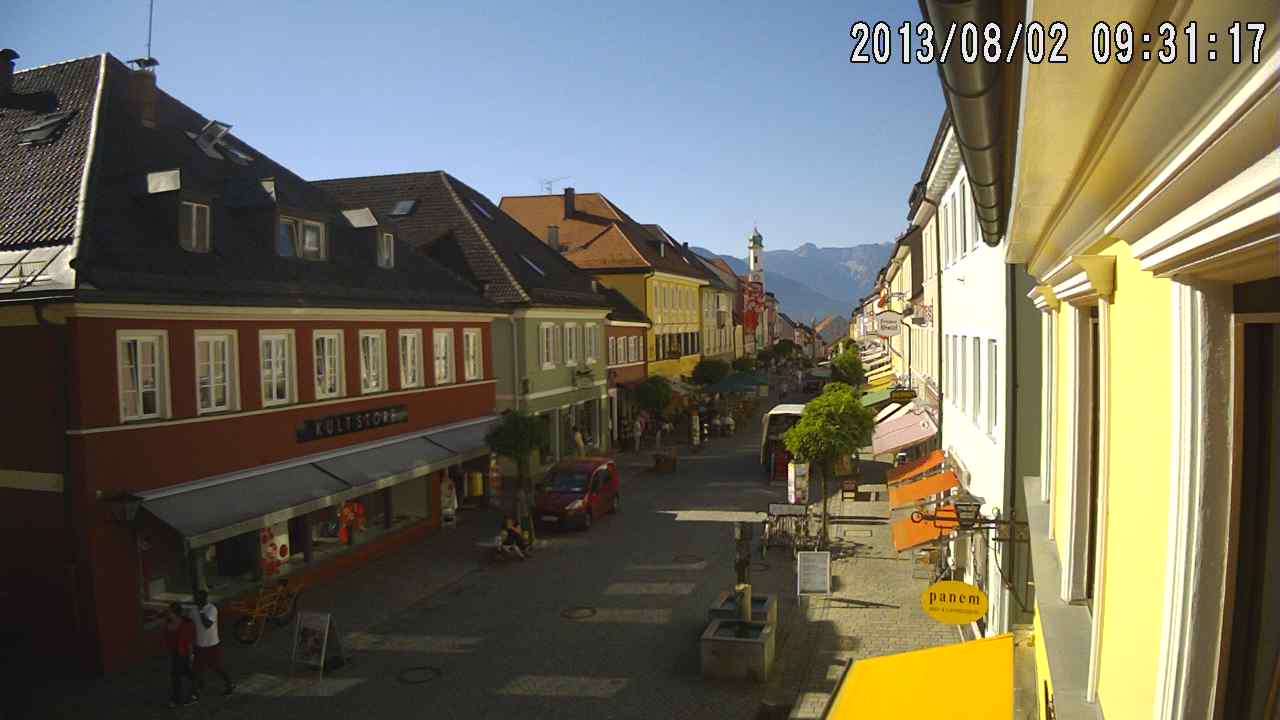} \\
          
          \includegraphics[width=.249\linewidth]{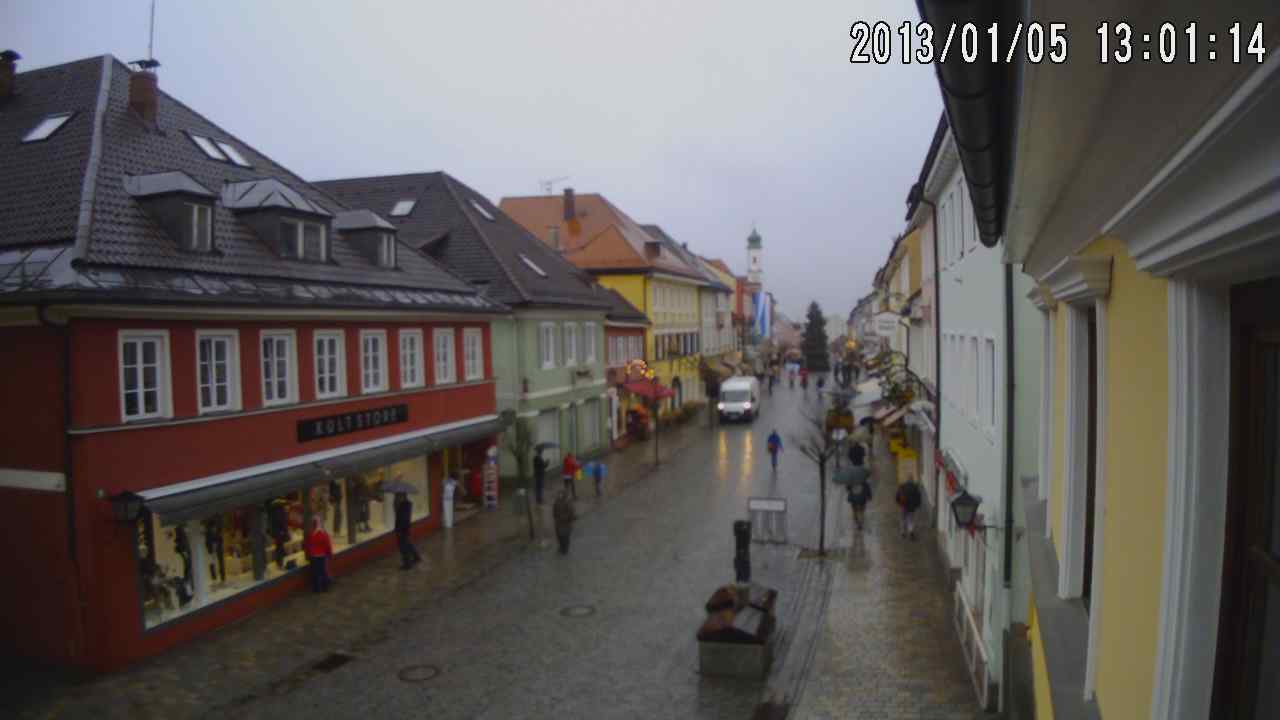} &
          \includegraphics[width=.249\linewidth]{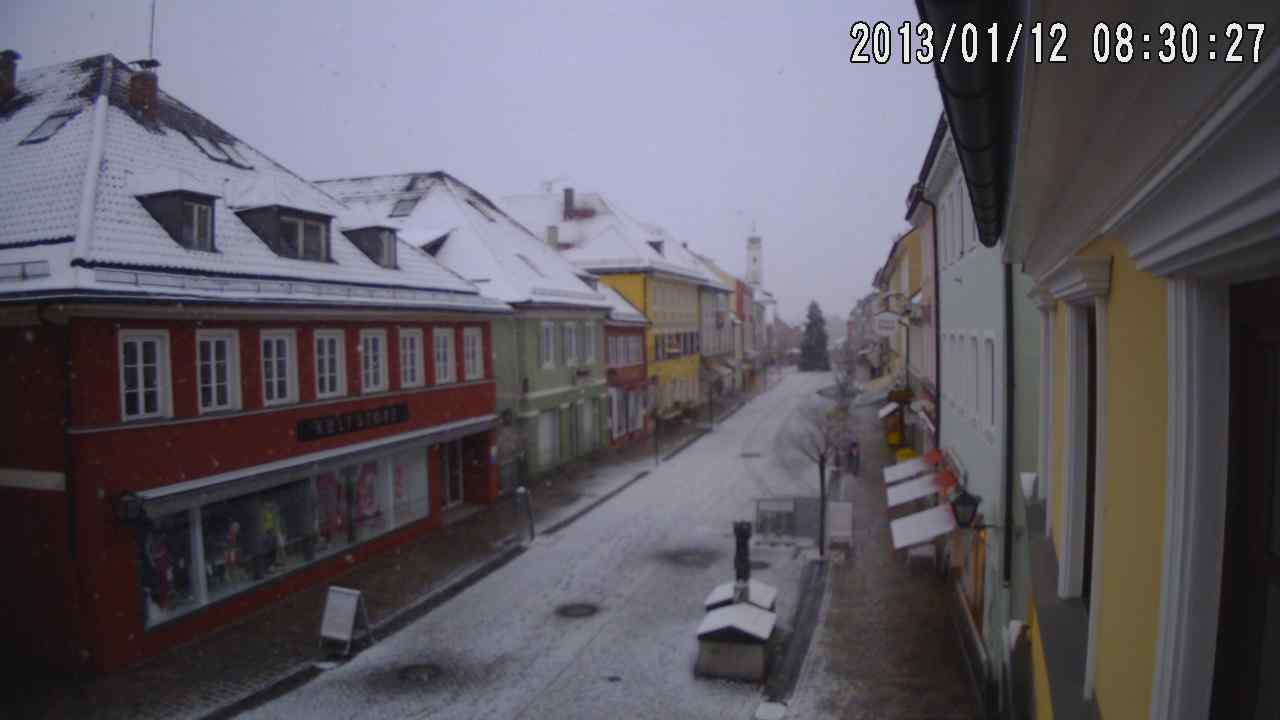} & 
          \includegraphics[width=.249\linewidth]{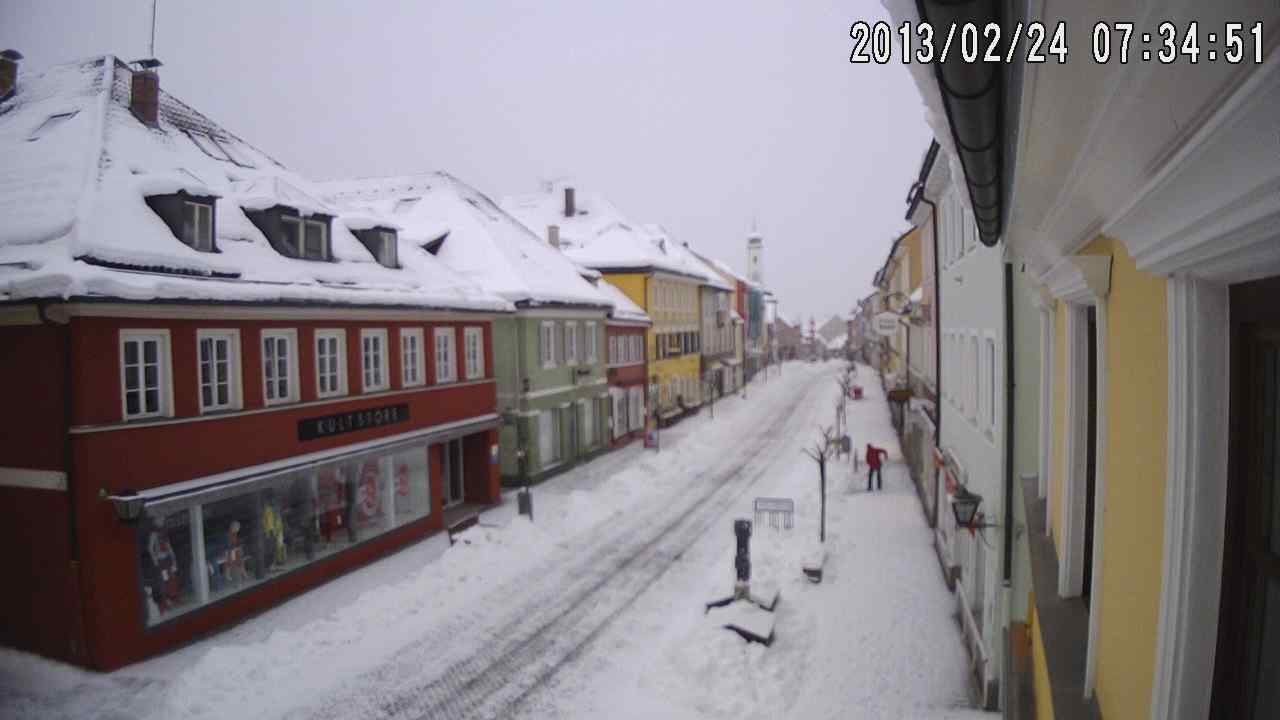} &
          \includegraphics[width=.249\linewidth]{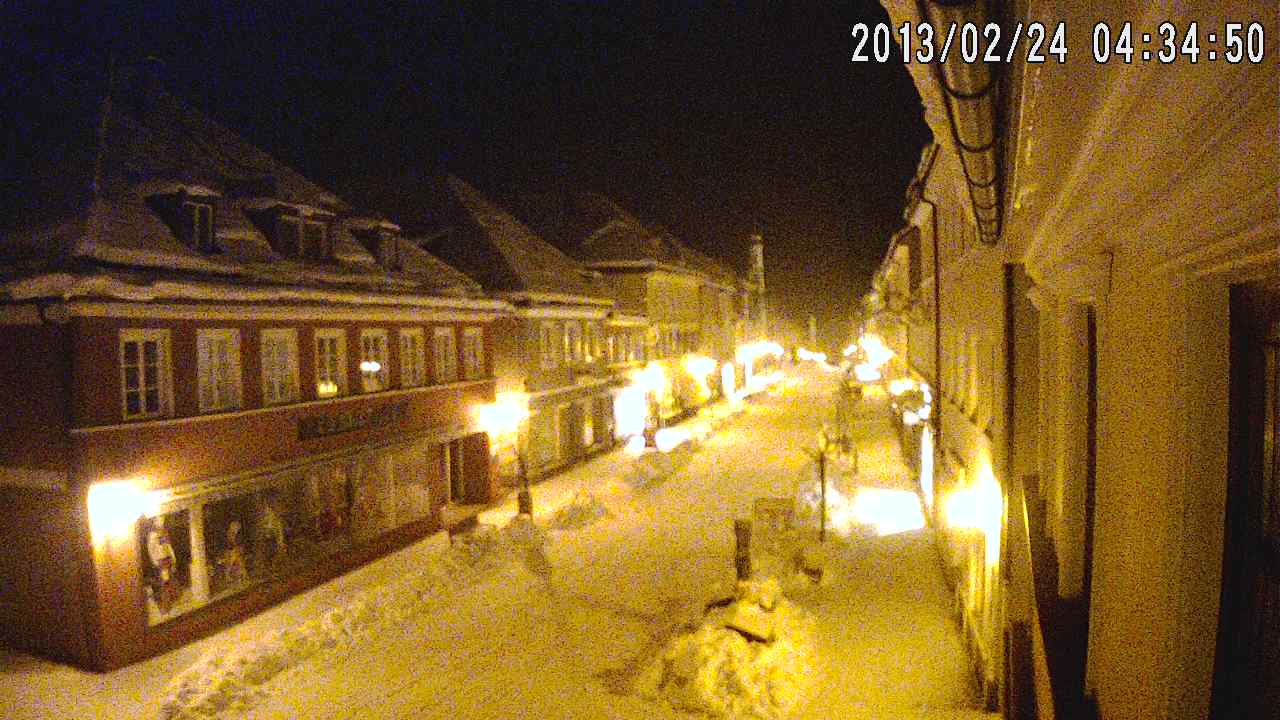} \\
      \end{tabular}
    \end{subfigure}
    \caption{An overhead image and the corresponding ground-level images from our CVT dataset.}
    \label{fig:dataset_sample}
\end{figure*}

\subsection{Image-driven mapping}
Typically image-based methods for generating maps start by extracting visual attributes from large-scale geotagged image collections and then apply a form of spatial smoothing, such as locally weighted averaging.
Examples include methods for mapping land cover~\cite{leung2010proximate}, scenicness~\cite{xie2011im2map}, snowfall~\cite{wang2013observing}, facial appearance~\cite{bessinger2016goes}, and a variety of other visual attributes~\cite{wang2016tracking}.

Integrating overhead imagery with image-driven mapping reduces the need for spatial smoothing, resulting in higher quality maps. This has been demonstrated for a variety of visual attributes, including building properties~\cite{workman2017unified}, natural beauty~\cite{workman2017beauty}, scene layouts~\cite{zhai2017crossview}, soundscapes~\cite{salem2018soundscape}, object distributions~\cite{greenwell2018goes, salem2019anything}, and land use~\cite{srivastava2019understanding}. 
Recent work has taken this to the extreme by synthesizing complete ground-level images~\cite{zhai2017crossview,deng2018like,regmi2018cross}. 

In this work, we perform image-driven mapping using overhead imagery, with location and time as additional context, resulting in high-resolution, dynamic maps of visual attributes. Most previous work has either ignored time, or merely used it to filter images outside of a time interval prior to spatial smoothing. Our work is similar to~\cite{workman2020dynamic}, but we focus on mapping visual attributes.

\subsection{Image geolocalization}
Recently, there has been a significant interest in the problem of image geolocalization, i.e., estimating the geographic location of the camera, or an object in the scene, given visual attributes extracted from the image~\cite{hays2008im2gps,weyand2016planet}. More recent work has shown that learning a feature mapping between ground-level and overhead image viewpoints enables image localization in regions without nearby ground-level images~\cite{lin2013cross,workman2015geocnn,lin2015learning,workman2015wide}. From this work, we see that image geolocalization requires the ability to extract visual attributes from ground-level images and an understanding of the geospatial distribution of these attributes. The former motivates our focus on generating high-quality, dynamic maps of visual attributes.

\subsection{Location context aids image understanding}
Studies have shown that additional context can aid visual understanding.  Tang et al.~\cite{tang2015improving} use the location an image was captured to improve classification accuracy. Luo et al.~\cite{luo2008event} use overhead imagery as additional context to improve event recognition in ground-level photos. Zhai et al.~\cite{zhai2018geotemporal} describe methods for learning image features using location and time metadata. Lee et al.~\cite{lee2015predicting} use map data to learn to estimate geo-informative attributes such as population density and elevation. Wang et al.~\cite{wang2016walk} use location information along with weather conditions to learn a feature representation for facial attribute classification. One potential use of our dynamic mapping approach would be as a model of the context needed for such image understanding applications.

\section{Cross-View Time (CVT) Dataset}

In an effort to support dynamic image-driven mapping, we introduce a new large-scale dataset that contains geotagged ground-level images, corresponding capture time, and co-located overhead images. We refer to our dataset as the Cross-View Time (CVT) dataset. It is similar to previous cross-view datasets~\cite{workman2015wide, workman2015geocnn,tian2017cross}, but ours is unique in providing timestamps for all images.

Our dataset is built from two sources of ground-level images. The first source is the Archive of Many Outdoor Scenes (AMOS)~\cite{jacobs07amos}, a collection of over a billion images captured from public outdoor webcams around the world. This subset~\cite{mihail2016sky} includes images captured between the years \num{2013} and \num{2014}, from \num{50} webcams, totaling \num{98633} images. Each image is associated with the location of the webcam and a timestamp (UTC) indicating when the image was captured. The second source is a subset of the Yahoo Flickr Creative Commons 100 Million Dataset (YFCC100M)~\cite{yfcc100m}. This subset~\cite{zhai2018geotemporal} contains geotagged outdoor images, with timestamps, captured by smartphones. 

We combined images from both of these sources to form a hybrid dataset containing \num{305011} ground-level images. For each image, we also downloaded an orthorectified overhead image from Bing Maps (\(800 \times 800\), \(0.60\) meters/pixel), centered on the geographic location. We randomly selected \num{25000} ground-level images, and the corresponding overhead images, and reserved them for testing. This resulted in a training dataset of \num{280011} image pairs. \figref{dataset_sample} shows example images from the CVT dataset.

\begin{figure}
  \centering
  \captionsetup[subfigure]{oneside,margin={1cm,1cm}}
  \includegraphics[width=\linewidth]{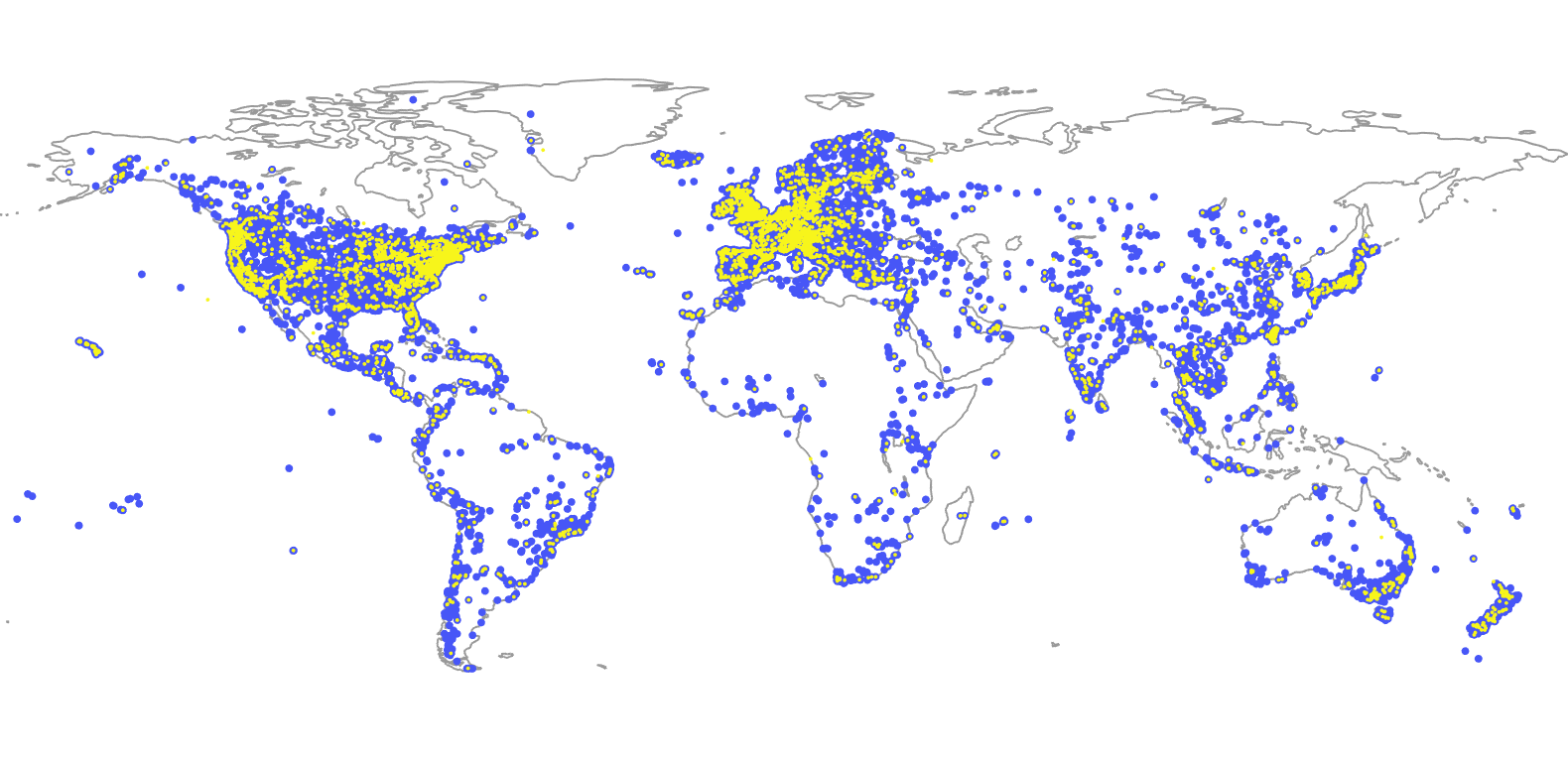}
  \caption{The spatial distribution of the CVT dataset. The blue (yellow) dots represent the training (testing) data.}
  \label{fig:data_dist}
\end{figure}

\begin{figure}
 
  \centering
  \includegraphics[width=\linewidth]{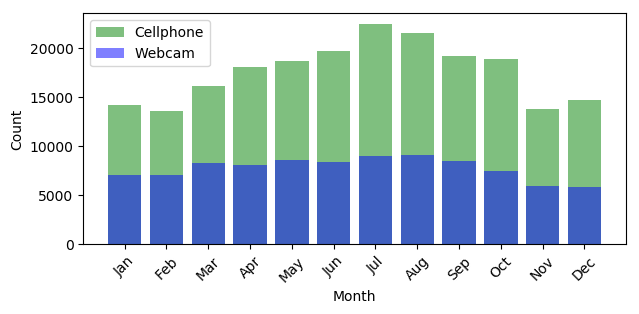}
  \includegraphics[width=\linewidth]{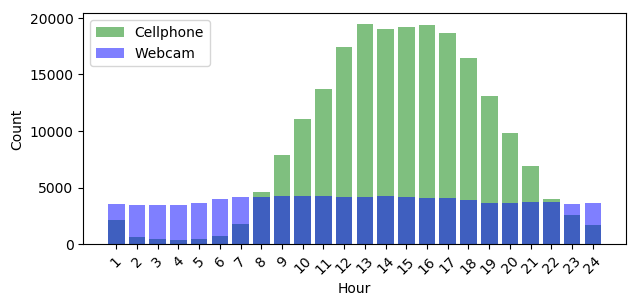}

  \caption{The temporal distribution of the CVT dataset.}
  \label{fig:time_dist}
\end{figure}

\figref{data_dist} shows the spatial distribution of the training images (blue dots) and testing images (yellow dots). Visual analysis of the distribution reveals that the images are captured from all over the world, with more images from Europe and the United States. Furthermore, examining the capture time associated with each image shows that the images cover a wide range of times. \figref{time_dist} visualizes the distribution over month and hour for both ground-level image sources. We observe that the webcam images are captured more uniformly across time than the cellphone images. The dataset is available at our project website.\punctfootnote{\url{https://tsalem.github.io/DynamicMaps/}}

\begin{figure*}
    \centering
    \includegraphics[width=\linewidth]{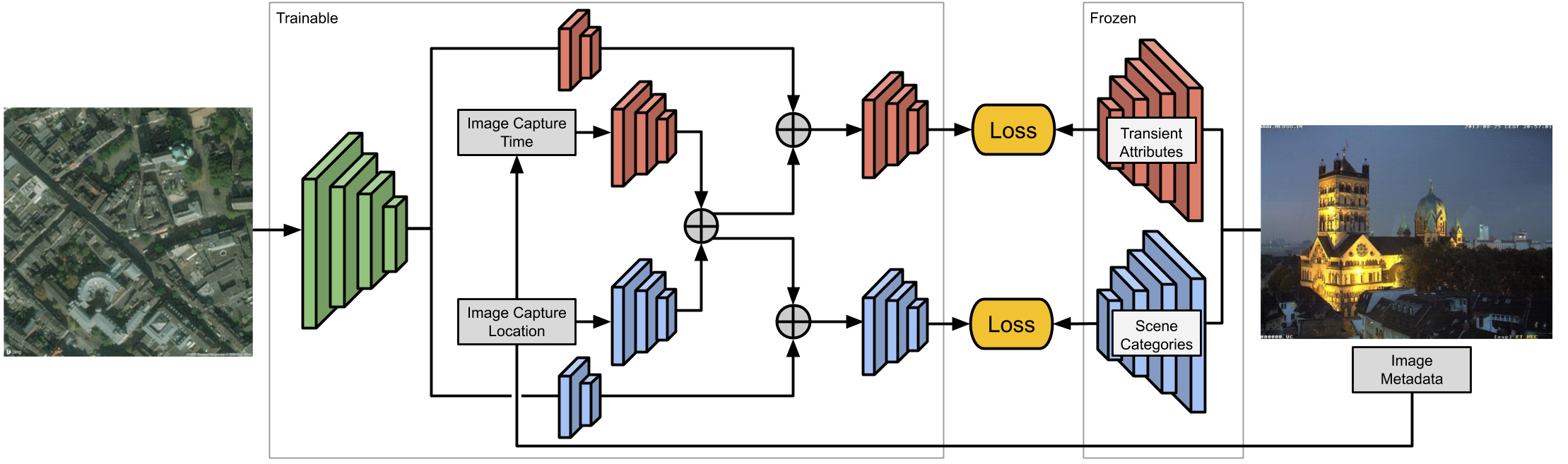}
    
    \caption{An overview of our network architecture, which includes the network we train to predict visual attributes (left) and the (frozen) networks we use to extract visual attributes from the ground-level images (right).}
    
    \label{fig:architecture}
    
\end{figure*}

\section{Dynamic Visual Appearance Mapping}

We present a general approach for dynamic visual appearance mapping that could be used to model a broad range of attributes and support many tasks.

\subsection{Problem Statement}

Our objective is to construct a map that represents the expected appearance at any geographic location and time. The expected appearance is defined using a set of visual attributes, which could be low level, such as a color histogram, or high level, such as the scene category. For a given visual attribute, \(a\), such a map can be modeled as a conditional probability distribution, \(P(a|t,l)\), given the time, \(t\), and location, \(l\), of the viewer. The distribution \(P(a|t,l)\) is challenging to learn because it essentially requires memorizing the Earth and how it changes over time.

We assume we are given a set of ground-level images, \(\{I_i\}\), each with associated capture time, \(\{t_i\}\), and geolocation metadata, \(\{l_i\}\). Furthermore, we assume we have the ability to calculate, or estimate with sufficient accuracy, each visual attribute from all images. The computed visual attributes, \(\{a_i\}\), can be considered samples from the probability distribution, \(P(a|t,l)\), and used for model fitting.

\subsection{Approach Overview}

To overcome the difficulty of directly modeling \(P(a|t,l)\), we also condition the distribution on an overhead image, \(I(l)\), of the location. Specifically, we define a conditional probability distribution, \(P(a|t,l,I(l))\). In doing so, the network no longer has to memorize, for example, the location of every road, river, and building in the world. We implement this using a mixture of convolutional and fully-connected neural networks to map from the conditioning variables to the parameters of distributions over a visual attribute, \(P(a | F(t,l,I(l);\Theta))\), where \(\Theta\) represents the parameters of all neural networks.

See \figref{architecture} for an overview of our complete architecture, which, in this case, simultaneously predicts two visual attributes. From the left, we first construct a feature embedding for each conditioning variable using a set of \emph{context} neural networks. We combine these context features to predict the visual attributes using a per-attribute, \emph{estimator} network. From the right, a set of pre-trained networks extract visual attributes from the ground-level images. These networks are only used for extracting visual attributes and are not trained in our framework.

This macro-architecture was carefully designed to balance several criteria. Most importantly, the overhead image is not dependent on time. This means that an overhead image is not required for every timestamp, \(t\), of interest. An overhead image {\em is} required for each location, but this is not a significant limitation given the wide availability of high-resolution satellite and aerial imagery. In addition, at inference time, feature extraction for the satellite image only needs to happen once, because the extraction process is not time or attribute dependent. 

\subsection{Network Architecture Details}

We propose a novel macro-architecture for modeling a dynamic visual appearance map. In this section, we define the specific neural network architectures and hyper-parameters we used for evaluation.

\textbf{Visual Attributes}
We focus on two visual attributes: \emph{Places}~\cite{zhou2017places}, which is a categorical distribution over \num{365} scene categories, and \emph{Transient}~\cite{laffont2014transient}, which is a multi-label attribute with \num{40} values that each reflect the degree of presence of different time-varying attributes, such as {\em sunny}, {\em cloudy}, or {\em gloomy}. To extract the \emph{Places} attributes, we use the pre-trained VGG-16~\cite{vgg} network. To extract the {\em Transient} attributes, we use a ResNet-50~\cite{he2016identity} model that we trained using the Transient Attributes Database~\cite{laffont2014transient}.

\textbf{Context Networks}
The context networks encode every conditioning variable, i.e., time, geographic location, and overhead image, to a \num{128}-dimensional feature vector. For the time and geolocation inputs, we use two similar encoding networks, each consisting of three fully connected layers with a ReLU activation. The layers have \num{256}, \num{512}, and \num{128} neurons respectively. The geographic location is represented in earth-centered earth-fixed coordinates, scaled to the range \([-1, 1]\). The time is factored into two components: the month of the year and the hour of the day. Each is scaled to the range \([-1, 1]\). For the overhead image, we use a ResNet-50 model to extract the \num{2048}-dimensional feature vector from the last global average pooling layer. This feature is passed to a per-attribute head. Each head consists of two fully connected layers that are randomly initialized using the Xavier scheme~\cite{glorot2010understanding}. The layers of each head have \num{256} and \num{128} neurons respectively, each with a ReLU activation.

\textbf{Estimator Networks}
For each visual attribute there is a separate estimator network, with only fully connected layers, that directly predicts the visual attribute. The input for these is the concatenation of the outputs of the context networks. For each estimator network, the first two layers (which have ReLU activations) contain $256$ and $512$ neurons, respectively. The third layer represents the output, with the number of neurons depending on the visual attribute. In this case, there are $365$ output neurons for the \emph{Places} estimator, with a \emph{softmax} activation, and $40$ for the \emph{Transient} estimator, with a \emph{sigmoid} activation.

\subsection{Implementation Details}

We jointly optimize all estimator and context networks with losses that reflect the quality of our prediction of the visual attributes extracted from ground-level images, \(\{I_i\}\). For the \emph{Places} estimator, the loss function is the KL divergence between attributes estimated from the ground-level image and the network output. For the \emph{Transient} estimator, the loss function is the mean squared error (MSE). These losses are optimized  using Adam~\cite{kingma2014adam} with mini-batches of size \num{32}. We applied $L_2$ regularization with scale \num{0.0005} and trained all models for \num{10} epochs with learning rate \num{0.001}.

All networks were implemented using TensorFlow~\cite{tensorflow} and will be shared with the community. Input images are resized to \(224 \times 224\) and scaled to $[\num{-1},1]$. We pre-trained the overhead context network to directly predict \emph{Places} and \emph{ImageNet} categories of co-located ground-level images, minimizing the KL divergence for each attribute. The weights are then frozen and only the added attribute-specific heads are trainable. 

For extracting \emph{Transient} attributes from the ground-level images, we train a ResNet-50 using the Transient Attributes Database~\cite{laffont2014transient} with the MSE loss. The weights were initialized randomly using the Xavier scheme, and optimized using Adam~\cite{kingma2014adam} until convergence with learning rate $0.001$ and batch size $64$. The resulting model achieves 3.04\% MSE on the test set, improving upon the 4.3\% MSE presented in the original work~\cite{laffont2014transient}. 
\begin{table}
    \centering
  
    \resizebox{\linewidth}{!}{  
        \begin{tabular}{ccccc}
            \toprule
            & \multicolumn{2}{c}{\emph{Places}} & \multicolumn{2}{c}{\emph{Transient}} \\
            \cmidrule(r){2-3}\cmidrule(l){4-5}
            Model & Top-1 & Top-5 & Within-\num{0.1} & Within-\num{0.2} \\
        \hline 
            {\em loc (k-NN)} & $17.68$ & $40.26$ & $50.96$ & $77.10$  \\
            {\em time (k-NN)} & $5.84$ & $17.82$ & $48.75$ & $75.77$  \\
            {\em time+loc (k-NN)} & $19.08$ & $41.15$ & $51.84$ & $77.51$ \\
        \hline
            {\em loc (CNN)} & $12.70$ & $32.45$ & $48.50$ & $75.45$  \\
            {\em time (CNN)} & $4.45$ & $16.91$ & $47.37$ & $75.34$  \\
            {\em time+loc (CNN)} & $17.05$ & $35.50$ & $54.69$ & $79.15$ \\
        \hline

            {\em sat (CNN)} & $15.16$ & $38.40$ & $49.87$ & $76.55$ \\
            {\em sat+loc (CNN)} & $16.98$ & $41.46$ & $50.57$ & $77.17$\\
            {\em sat+time (CNN)} & $19.66$ & $40.78$ & $56.14$ & $79.79$\\
            \hline
            {\em \textbf{sat+time+loc (CNN)}} & \textbf{21.58} & \textbf{44.00} 
            & \textbf{56.91} & \textbf{80.55} \\
            
            \bottomrule
        \end{tabular}
    }
    
    \caption{A comparison of the prediction accuracy of our full approach, \emph{sat+time+loc}, with various baseline models.}
    
    \label{tbl:accuracy}
    
\end{table}

\section{Evaluation}

We evaluate our approach using the CVT dataset quantitatively, qualitatively, and on a variety of applications. We use Top-1 and Top-5 classification accuracy as the metric for evaluating quality of the \emph{Places} attribute predictions. For the \emph{Transient} attribute we use the percent of attribute predictions within a threshold (0.1 or 0.2) of the ground truth. In both cases, these are averaged across the full test set.

\begin{figure*}

  \centering

  \setlength\tabcolsep{1pt}

  \begin{tabular}{lccc}
    
    & \emph{lush} & \emph{warm} & \emph{gloomy} \\    
        
    \raisebox{.4\height}{\rotatebox{90}{\em January}} &
    \includegraphics[width=.32\linewidth]{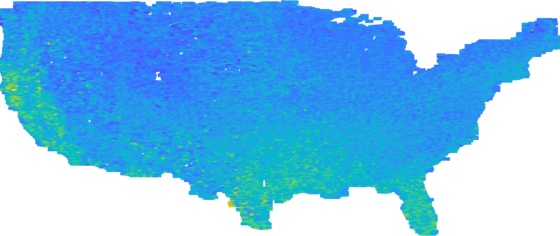} &
    \includegraphics[width=.32\linewidth]{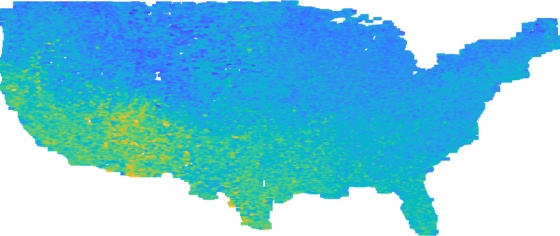} &
    \includegraphics[width=.32\linewidth]{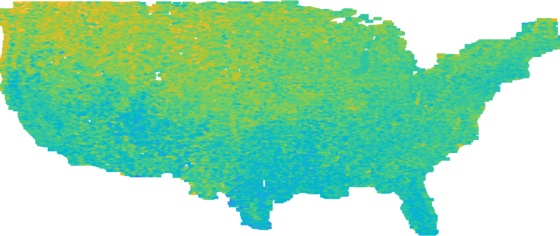} \\ 
    
    \raisebox{1.1\height}{\rotatebox{90}{\em April}} &
    \includegraphics[width=.32\linewidth]{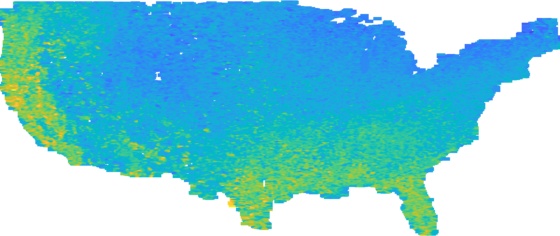} &
    \includegraphics[width=.32\linewidth]{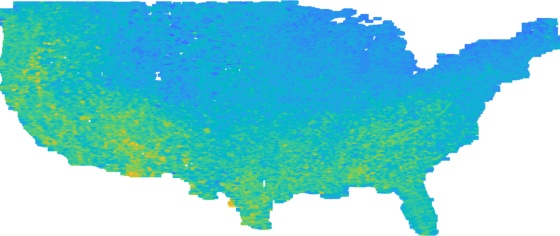} &
    \includegraphics[width=.32\linewidth]{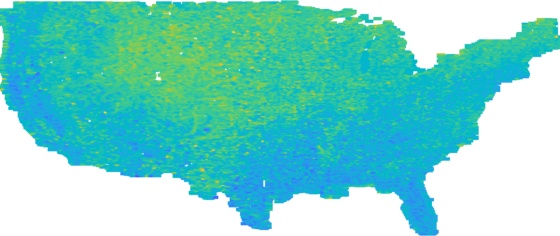} \\
    
    \raisebox{1.5\height}{\rotatebox{90}{\em July}} &
    \includegraphics[width=.32\linewidth]{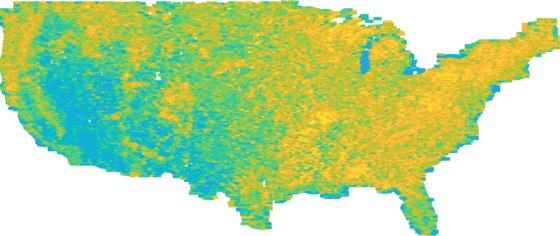} &
    \includegraphics[width=.32\linewidth]{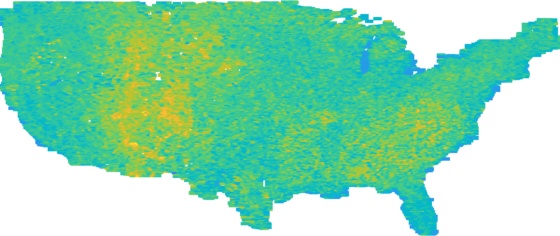} &
    \includegraphics[width=.32\linewidth]{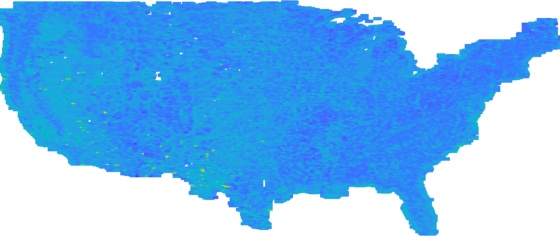} \\
        
  \end{tabular}

  \caption{Dynamic visual attribute maps for different {\em Transient} attributes. In each, yellow (blue) corresponds to a higher (lower) value for the corresponding attribute. Each attribute exhibits unique spatial and temporal patterns, which closely match the authors' personal travel experiences.}

  \label{fig:maps}

\end{figure*}

\subsection{Exploratory Dataset Analysis}

To better understand the relationship between location, time, and these attributes, we conducted a preliminary study without using overhead imagery. For the \emph{Places} attribute, we use a k-NN classifier ($k=30$) to explore this relationship. As features we used time (linear) and latitude/longitude (degrees). We scaled the time using grid-search to optimize the accuracy when using all features. The resulting classifier obtained 19.08\% accuracy on the test set (see Table~\ref{tbl:accuracy}). If we remove the time feature, the accuracy drops a small amount to 17.68\%. If we remove both location features, the accuracy is 5.84\%, which is better than ignoring all features (1.96\%). From this, we can see that the \emph{Places} attribute is highly dependent on location but less-so on time. We were surprised that the time feature by itself resulted in such high accuracy. We suspect that this is due to differences in the types of pictures taken at different times of year.

For the \emph{Transient} attributes, we used a similar setup. The only change was using a k-NN regression model. Table~\ref{tbl:accuracy} shows that the difference between features is less dramatic than it was for the \emph{Places} attributes. Instead, we focus on the impact of removing the location and time features on the individual attributes. When removing the location feature, we found, for example, that the accuracy for some attributes went down more than 6\% (e.g., \emph{busy}, \emph{fog}, \emph{gloomy}) while for others it went up more than 2\% (e.g., \emph{dawndusk}, \emph{dark}, \emph{night}). For the time feature, we found that the accuracy went down for all attributes, with some going down significantly (e.g., \emph{winter}, \emph{snow}, \emph{lush}) but others only marginally (e.g., \emph{rain}, \emph{sunrisesunset}, \emph{sentimental}). 

These results highlight that the relationship between visual attributes, location, and time is complex and that our dataset enables us to translate intuitive notions into concrete experimental results. 

\subsection{Quantitative Evaluation}

We trained several variants of our full model, \emph{sat+time+loc}. For each, we omit either one or two of the conditioning variables but retain all other aspects. We use the same training data, training approach, and micro-architectures. In total, we trained six baseline models: \emph{loc}, \emph{time}, \emph{sat}, \emph{time+loc}, \emph{sat+loc}, and \emph{sat+time}. We evaluate the accuracy of all methods on the test set.

\tblref{accuracy} shows the accuracy for all approaches on both visual attributes. We find that our method has the highest accuracy. However, the ranking of baseline models changes depending on the visual attribute. For example, the accuracy for the \emph{sat+loc} model is relatively worse for the \emph{Transient} attribute than the \emph{Places} attribute. This makes sense because the former is highly dependent on when an image was captured and the latter is more stable over time. We also note the significant improvement, for both attributes, obtained by including overhead imagery in the model. For example, the \emph{time+loc} model is significantly worse than our full model.

\subsection{Examples of Visual Attribute Maps}

\figref{maps} shows several example attribute maps rendered from our model. To construct these we use the CVUSA dataset~\cite{workman2015wide}, which contains overhead imagery across the continental United States. Specifically, we use a subset of \num{488243} overhead images associated with the Flickr images in the dataset. For each overhead image, we compute visual attributes using our full model, \emph{sat+time+loc}. We specify the time of day as 4pm, and vary the month. 
 
The trends we observe are in line with our expectations. For example, for the transient attribute \emph{lush}, which refers to vegetation growing, January has low values (blue) in the northernmost regions. However, the highest estimates (yellow) include regions like Florida and California. The lushness estimate progressively increases from January through April, achieving its highest value in July. Similarly, the \emph{warm} attribute is highest in the southwest during both winter and spring, but reaches higher overall values in the summer months. Meanwhile, the \emph{gloomy} attribute is highest during winter, with a bias towards the Pacific Northwest, and decreases during the summer. 

\begin{figure}
    \begin{subfigure}{0.15\textwidth}
         \begin{tabular}{c}
        \includegraphics[scale=.3]{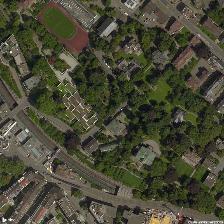}\\
        {\emph{Latitude}: $47.367$}\\ 
        {\emph{Longitude}: $8.55$}
        \end{tabular}
    \end{subfigure}%
    \hfill
    \begin{tabular}{c}
    \begin{subfigure}[t]{0.85\textwidth}
       \includegraphics[scale=.2]{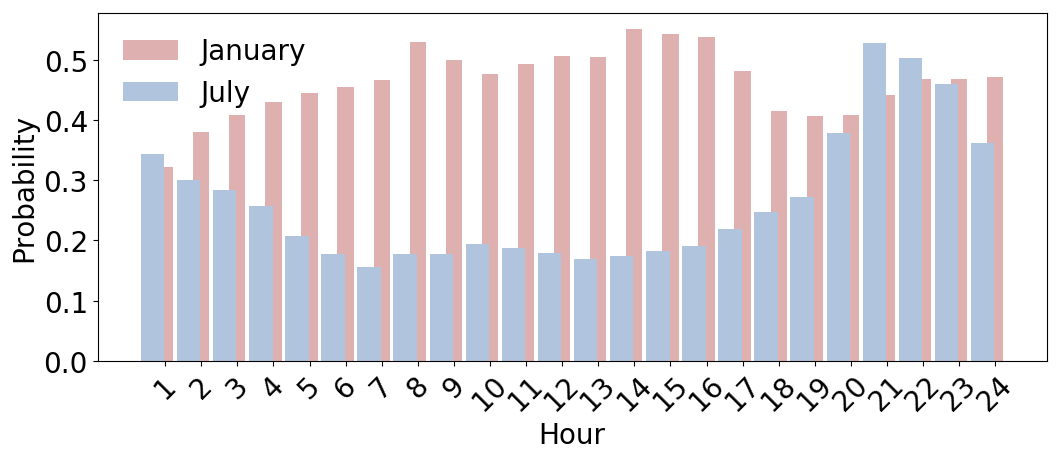}
     \end{subfigure}
      \\
      \begin{subfigure}[t]{0.85\textwidth}
      \includegraphics[scale=.2]{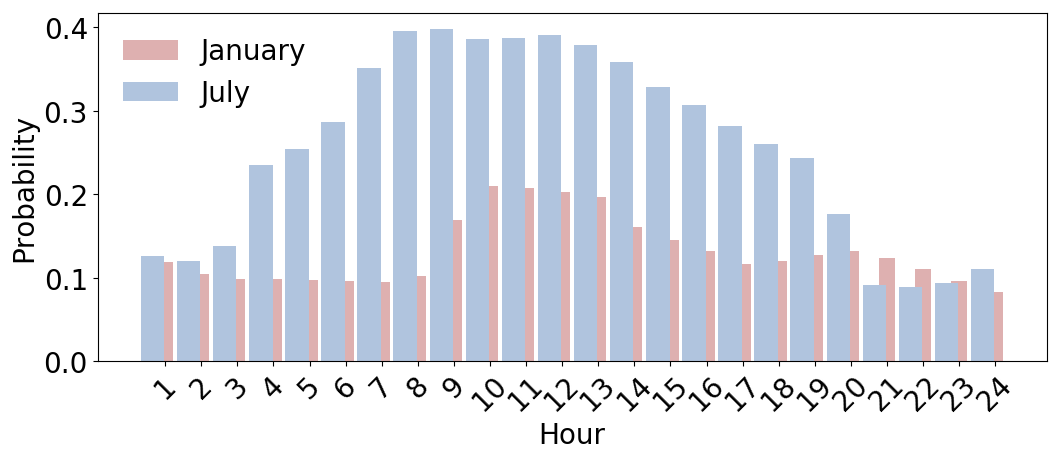}
      \end{subfigure}
      \end{tabular}
    \caption{For a given location and corresponding overhead image, (top, right) shows the predictions from our model for the \emph{cold} attribute. Similarly, (bottom, right) shows the warm attribute. Both examples show that our model has learned dynamic patterns of visual attributes.}
    \label{fig:over_hours}
\end{figure}

\figref{over_hours} shows an example of how the estimated attribute varies over time. Our proposed model captures changes in the different attributes not only over months of the year but also over hours of the day. In \figref{over_hours} (top, right) the \emph{cold} attribute during a day in January is higher than a day in July, whereas in \figref{over_hours} (bottom, right) the \emph{warm} attribute is opposite. These results demonstrate that our model has captured temporal trends. 

\section{Applications}

We show how our dynamic mapping approach can be used to support three image-understanding applications: localization, retrieval, and metadata verification. Together, they demonstrate that combining overhead imagery, location, and time is critical for correctly modeling the dynamic distribution of visual attributes.

A key component of each application is computing the distance between the visual attributes of a ground-level image and the visual attributes predicted by our model. For the {\em Places} attribute we use the KL divergence and for the {\em Transient} attribute we use the $L_2$ distance. We also define {\em Combine} which is a weighted average of these two, with $\lambda$ as the weight for {\em Places} and $1-\lambda$ for {\em Transient}. The value of $\lambda$ is selected empirically for each application.

\begin{table}

    \resizebox{\linewidth}{!}{  
        \begin{tabular}{ccccccc}
            \toprule
            & \multicolumn{2}{c}{\emph{Transient}} & \multicolumn{2}{c}{\emph{Places}} & \multicolumn{2}{c}{\emph{Combine}} \\
              \cmidrule(r){2-3}\cmidrule(l){4-5} \cmidrule(l){6-7}
            Context & Top-1\% & Top-5\% & Top-1\% & Top-5\% & Top-1\% & Top-5\% \\
            \hline
            {\em sat} & $4.80$ & $15.30$ & $18.80$ & $42.00$ & $21.60$ & $42.60$\\
            {\em sat+loc} & $5.50$ & $15.40$ & $23.00$ & $45.40$ & $23.90$ & $45.00$\\
            {\em sat+time} &  $13.10$ & $22.50$ & $23.90$ & $43.60$ & $24.90$ & $44.00$\\
            {\em \textbf{sat+time+loc}} & $13.70$ & $25.00$ & $28.70$ & $47.60$ & \textbf{31.20} &\textbf{49.30} \\
            \bottomrule
        \end{tabular}
    }
    
    \centering
    \caption{Localization accuracy of different models and distance measures.}
    
    \label{tbl:loc_accuracy}
    
\end{table}

\subsection{Application: Image Localization}
\label{loc_task}

We evaluated the accuracy of our models on the task of image geolocalization, using a set of \num{1000} ground-level query images randomly sampled from the test set. To localize an image, we first extract its visual attributes. Then, we predict the visual attributes for all \num{1000} overhead images. As context, we use the location of the corresponding overhead image and the capture time of the ground-level image. We compute the distance between these predicted attributes and the attributes extracted from the image. We use $\lambda=0.58$ when computing the {\em Combine} distance.

Table \ref{tbl:loc_accuracy} shows the results of this experiment. Each number represents the percentage of query images that were correctly localized within the Top-k\% of candidate locations. For a given threshold, a higher percentage localized is better. This experiment shows that our full model outperforms the baselines and that using the {\em Combine} distance results in the highest accuracy. It also shows that the time attribute is essential when localizing using the {\em Transient} feature. In all cases, using only the imagery, which is the current state of the art, results in the lowest accuracy.

\begin{figure*}

  \centering

  \setlength\tabcolsep{1pt}

  \begin{tabular}{ccc}
    Overhead Image  & 5pm (UTC) & 2am (UTC) \\    
    \includegraphics[scale=0.42]{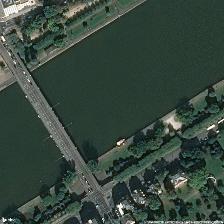} &\hspace{2mm}
    \includegraphics[width=0.29\textwidth]{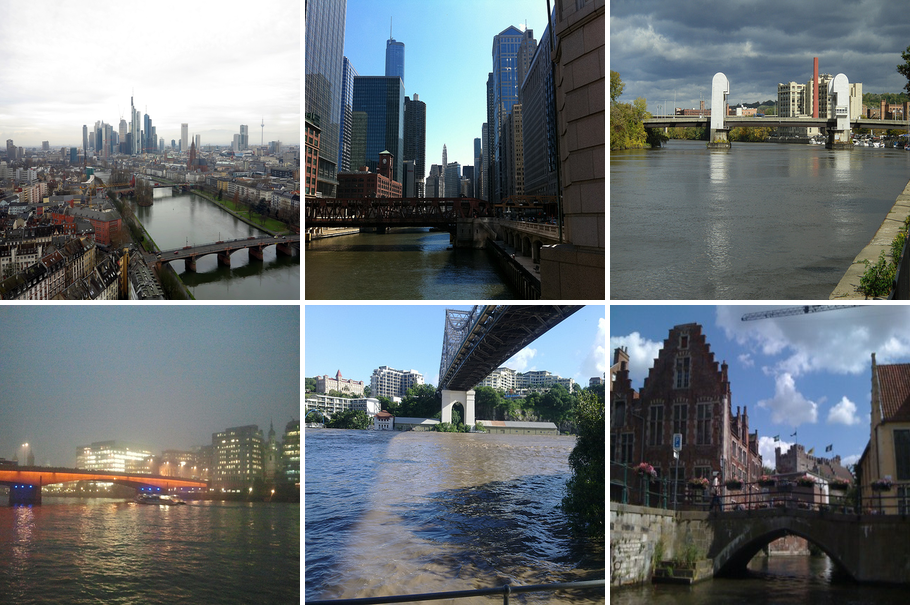} &\hspace{2mm}
    \includegraphics[width=0.29\textwidth]{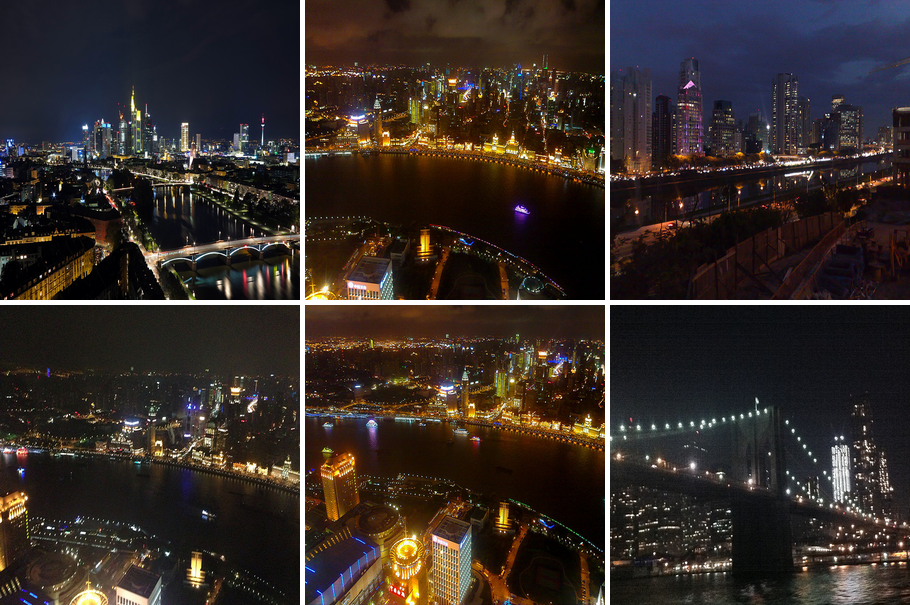}\\

    \includegraphics[scale=0.42]{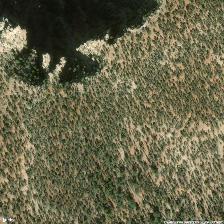} &\hspace{2mm}
    \includegraphics[width=0.29\textwidth]{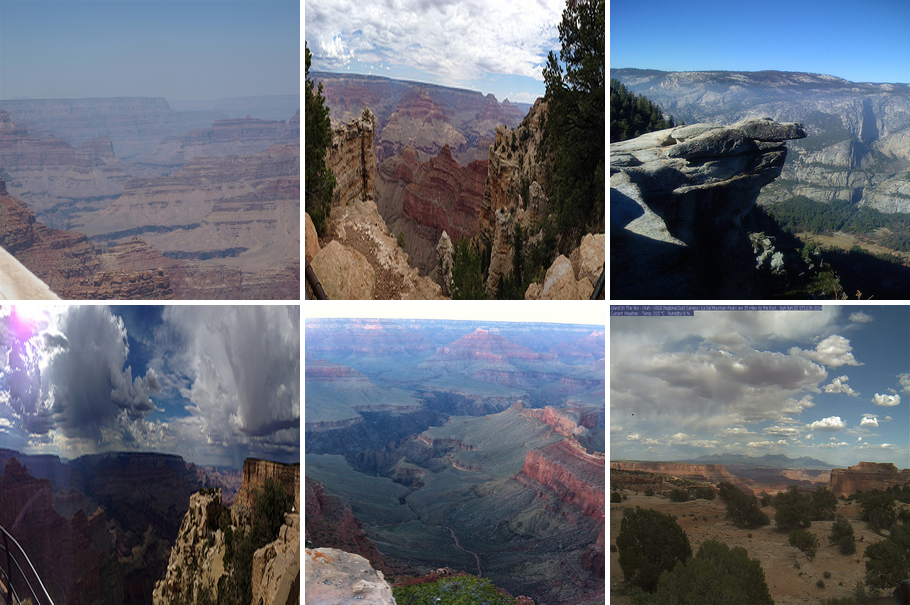} &\hspace{2mm}
    \includegraphics[width=0.29\textwidth]{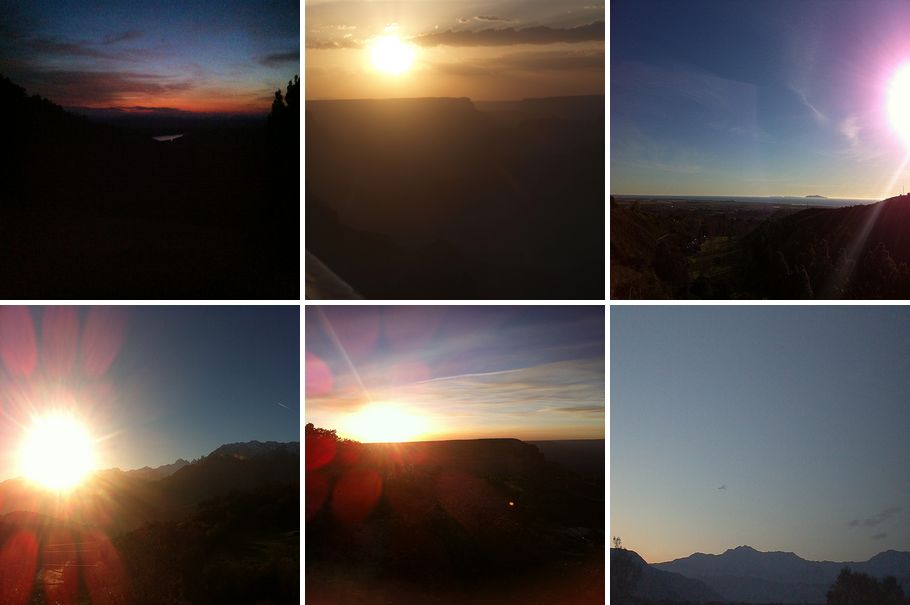}
        
  \end{tabular}
  
  \caption{For each overhead image, we predict the visual attributes using our full model and compute the average distance between them and those of the ground-level images in the test set. (left) The overhead images of two query locations. The closest images when using August at 5pm as input (middle) and when using August at 2am (right).}

  \label{fig:matching}

\end{figure*}

\subsection{Application: Image Retrieval}

In this qualitative application, we show how we can use our model to retrieve a set of ground-level images that would be likely to be observed at a given location and time. We start with an overhead image, specify a time of interest, and predict the visual attributes. We use the {\em Combine} distance defined in the previous section to find the closest ground-level images. In \figref{matching}, we show examples of images retrieved using this process. We observe that the ground-level images contain the expected scene type and appear to be from the appropriate time of day. For example, the top left overhead image contains a bridge and the closest ground-level images are visually consistent at both input timestamps. 

\subsection{Application: Metadata Verification}

\begin{figure*}
    \centering
    \hspace{0.3cm}
    \raisebox{-0.5\height}{\includegraphics[height=3.5cm,width=3.7cm]{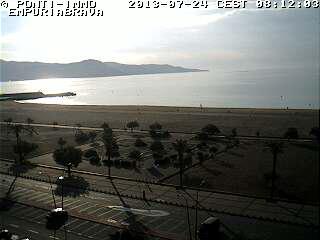}}
    \raisebox{-0.5\height}{\includegraphics[height=3.5cm,width=3.7cm]{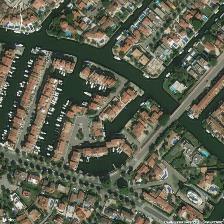}}
    \hspace{1.1cm}
    \centering
    \raisebox{-0.5\height}{\includegraphics[height=3.5cm,width=3.7cm]{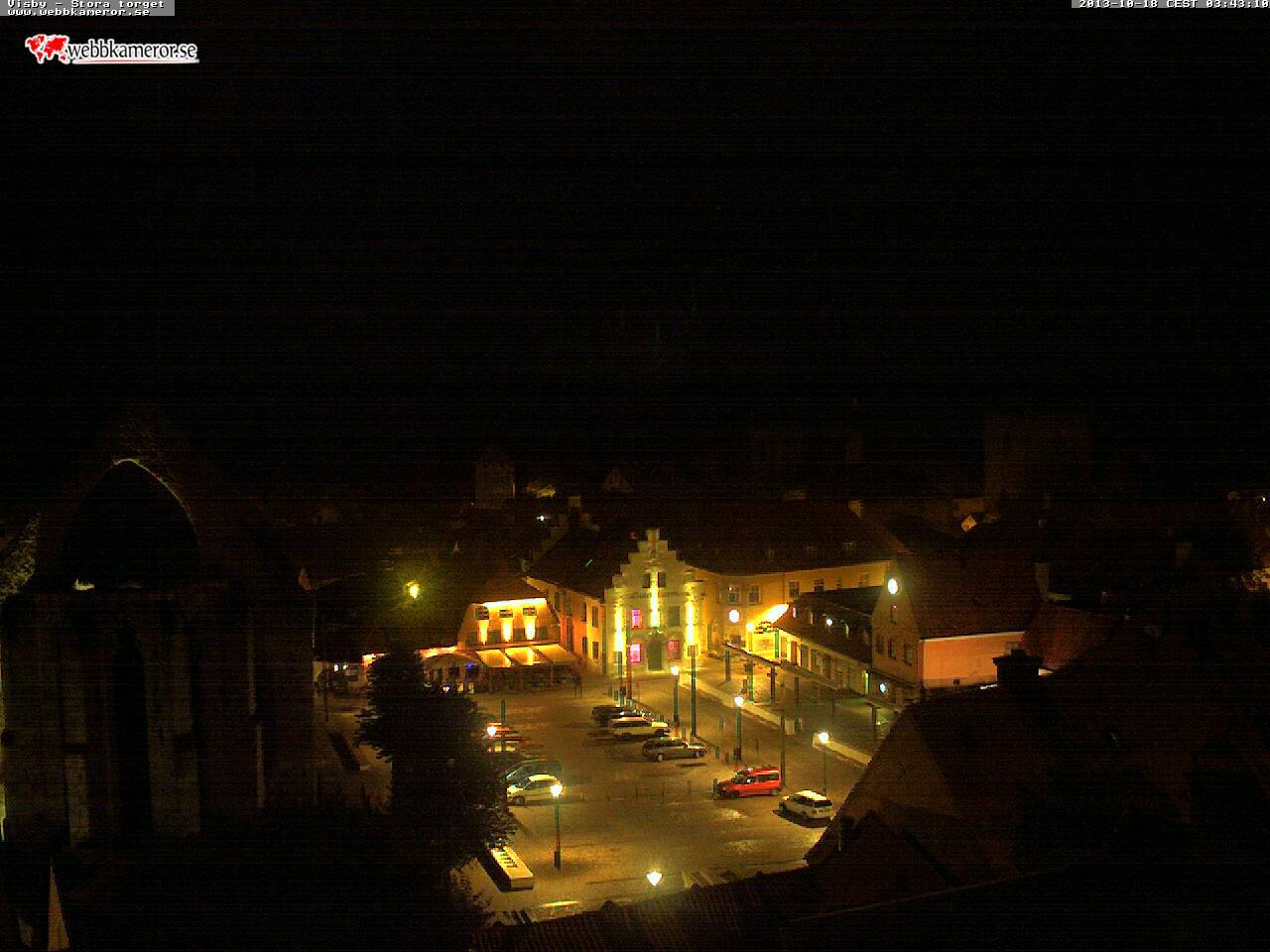}}
    \raisebox{-0.5\height}{\includegraphics[height=3.5cm,width=3.7cm]{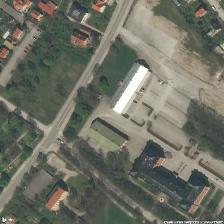}}
    
    \raisebox{-0.5\height}{\includegraphics[scale = 0.3]{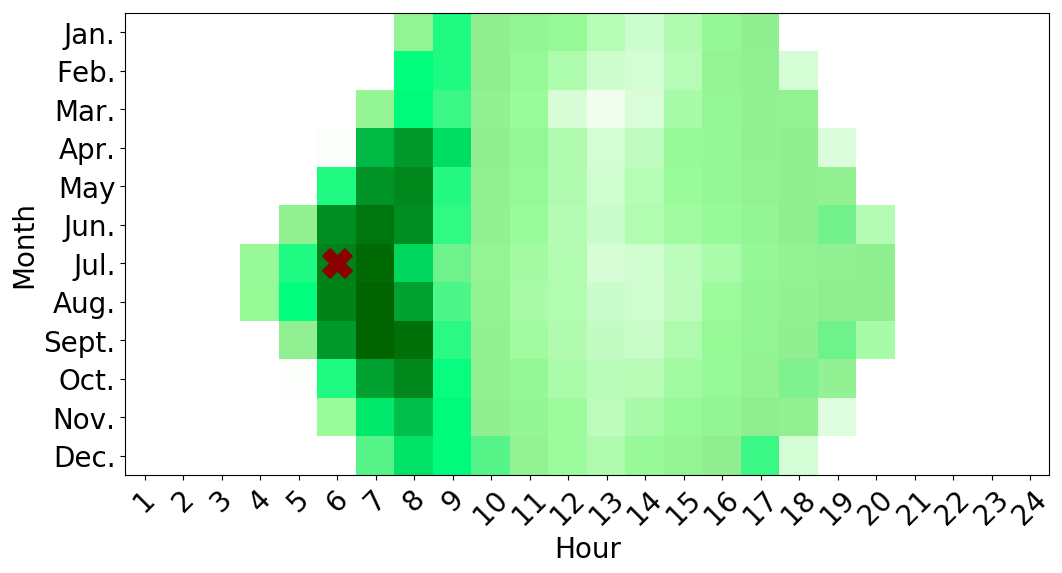}}
    \hspace{.5cm}
    \raisebox{-0.5\height}{\includegraphics[scale = 0.3]{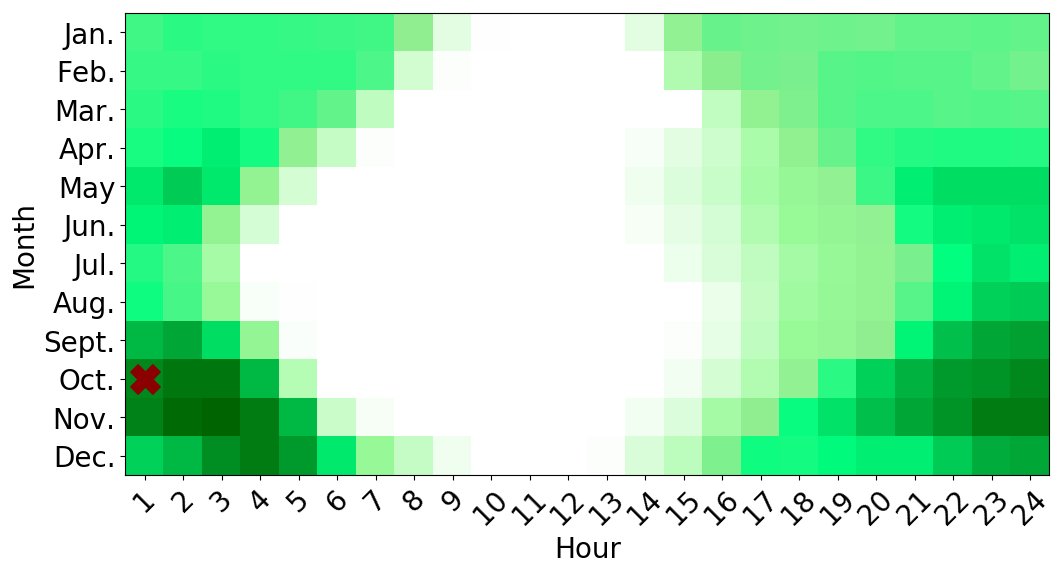}}
   
    \caption{Two examples highlighting temporal patterns learned by our model. For each example, we show the original image and the overhead image of its location. For every possible hour and month, we use our full model to predict the visual attributes. The heatmap shows the distance between the true and predicted visual attributes, with dark green (white) representing smaller (larger) distances.
    }
  \label{fig:forensics} 
  
\end{figure*}

We focus on verifying the time that an image, with known location, was captured. For a given ground-level image, we first extract its visual attributes and then predict the visual attributes for a range of different times. We compute the distance between the actual and predicted attributes resulting in a distance for each possible time. \figref{forensics} shows heatmaps of these distances for two test examples, using our full model and the \emph{Combine} distance. These show that our model is able to identify a small set of likely times.

We conducted a quantitative evaluation on a sample of \num{2000} images. For each image, we compute the distances as described above and then rank the times based on distance. Ideally, the correct time will have the lowest distance. In Table \ref{tbl:metadata_accuracy}, 
we show the percent of images for which the correct time was within the Top-k\% of possible times. The results show that the \emph{Combine} distance outperforms both \emph{Places} and \emph{Transient}. While this approach does not fully solve the problem of detecting metadata falsification, it demonstrates that our model could be an important part of the solution.

\begin{table}

    \resizebox{\linewidth}{!}{  
        \begin{tabular}{ccccccc}
            \toprule
            & \multicolumn{2}{c}{\emph{Transient}} & \multicolumn{2}{c}{\emph{Places}} & \multicolumn{2}{c}{\emph{Combine}} \\
              \cmidrule(r){2-3}\cmidrule(l){4-5} \cmidrule(l){6-7}
            Context & Top-1\% & Top-5\% & Top-1\% & Top-5\% & Top-1\% & Top-5\% \\
            \hline
            {\em time} &  $13.4$ & $48.90$ & $10.85$ & $45.40$ & $13.25$ & $48.85$\\
            {\em loc+time} &  $31.50$ & $81.50$ & $27.20$ & $74.90$ & $36.20$ & $82.10$\\
            {\em sat+time} &  $34.50$ & $81.65$ & $31.65$ & $79.55$ & $37.50$ & $83.30$\\
            {\em \textbf{sat+time+loc}} & $32.95$ & $82.30$ & $33.60$ & $79.85$ & \textbf{40.30} &\textbf{84.35} \\
            \bottomrule
        \end{tabular}
    }
    
    \centering
    \caption{Time verification accuracy of various baselines and two thresholds. Our approach with the \emph{Combine} distance outperforms all other methods.}
    
    \label{tbl:metadata_accuracy}
\end{table}

\section{Conclusion}

We introduced a novel method for constructing dynamic visual attribute maps. In several large scale experiments, we demonstrated the practical utility of the model and highlighted the importance of including time, location, and an overhead image of the location as conditioning variables. Such a model has many potential uses, including image-driven mapping, image localization, and metadata verification. In future work, we plan to focus on adapting this model to more directly support the application of metadata verification and to include additional visual attributes.

\ifcvprfinal
{\noindent\textbf{Acknowledgements:}
We gratefully acknowledge the financial support of an NSF CAREER grant (IIS-1553116), the University of Kentucky Center for Computational Sciences, and a Google Faculty Research Award. Thanks to Armin Hadzic for helpful feedback on the manuscript.
}
\fi

{\small
\bibliographystyle{ieee_fullname}
\bibliography{biblio}

\begin{thebibliography}{10}\itemsep=-1pt

\bibitem{tensorflow}
M. Abadi~et al.
\newblock Tensorflow: A system for large-scale machine learning.
\newblock In {\em {USENIX} Symposium on Operating Systems Design and
  Implementation}, 2016.

\bibitem{bessinger2016goes}
Zachary Bessinger, Chris Stauffer, and Nathan Jacobs.
\newblock Who goes there? {A}pproaches to mapping facial appearance diversity.
\newblock In {\em ACM SIGSPATIAL International Conference on Advances in
  Geographic Information Systems}, 2016.

\bibitem{bharati2019beyond}
Aparna Bharati, Daniel Moreira, Joel Brogan, Patricia Hale, Kevin Bowyer,
  Patrick Flynn, Anderson Rocha, and Walter Scheirer.
\newblock Beyond pixels: Image provenance analysis leveraging metadata.
\newblock In {\em IEEE Winter Conference on Applications of Computer Vision},
  2019.

\bibitem{bianchi2012image}
Tiziano Bianchi and Alessandro Piva.
\newblock Image forgery localization via block-grained analysis of jpeg
  artifacts.
\newblock {\em IEEE Transactions on Information Forensics and Security},
  7(3):1003--1017, 2012.

\bibitem{deng2018like}
Xueqing Deng, Yi Zhu, and Shawn Newsam.
\newblock What is it like down there? {G}enerating dense ground-level views and
  image features from overhead imagery using conditional generative adversarial
  networks.
\newblock In {\em ACM SIGSPATIAL International Conference on Advances in
  Geographic Information Systems}, 2018.

\bibitem{farid2009image}
Hany Farid.
\newblock Image forgery detection.
\newblock {\em IEEE Signal Processing Magazine}, 26(2):16--25, 2009.

\bibitem{gebru2017using}
Timnit Gebru, Jonathan Krause, Yilun Wang, Duyun Chen, Jia Deng, Erez~Lieberman
  Aiden, and Li Fei-Fei.
\newblock Using deep learning and google street view to estimate the
  demographic makeup of neighborhoods across the united states.
\newblock {\em Proceedings of the National Academy of Sciences},
  114(50):13108--13113, 2017.

\bibitem{glorot2010understanding}
Xavier Glorot and Yoshua Bengio.
\newblock Understanding the difficulty of training deep feedforward neural
  networks.
\newblock In {\em International Conference on Artificial Intelligence and
  Statistics}, 2010.

\bibitem{greenwell2018goes}
Connor Greenwell, Scott Workman, and Nathan Jacobs.
\newblock What goes where: Predicting object distributions from above.
\newblock In {\em IEEE International Geoscience and Remote Sensing Symposium},
  2018.

\bibitem{guan2019mfc}
Haiying Guan, Mark Kozak, Eric Robertson, Yooyoung Lee, Amy~N Yates, Andrew
  Delgado, Daniel Zhou, Timothee Kheyrkhah, Jeff Smith, and Jonathan Fiscus.
\newblock {MFC} datasets: Large-scale benchmark datasets for media forensic
  challenge evaluation.
\newblock In {\em IEEE Winter Conference on Applications of Computer Vision},
  2019.

\bibitem{hays2008im2gps}
James Hays and Alexei~A Efros.
\newblock {IM2GPS}: Estimating geographic information from a single image.
\newblock In {\em IEEE Conference on Computer Vision and Pattern Recognition},
  2008.

\bibitem{he2016identity}
Kaiming He, Xiangyu Zhang, Shaoqing Ren, and Jian Sun.
\newblock Identity mappings in deep residual networks.
\newblock In {\em European Conference on Computer Vision}, 2016.

\bibitem{jacobs07amos}
Nathan Jacobs, Nathaniel Roman, and Robert Pless.
\newblock Consistent temporal variations in many outdoor scenes.
\newblock In {\em IEEE Conference on Computer Vision and Pattern Recognition},
  2007.

\bibitem{kingma2014adam}
Diederik Kingma and Jimmy Ba.
\newblock Adam: A method for stochastic optimization.
\newblock In {\em International Conference on Learning Representations}, 2014.

\bibitem{laffont2014transient}
Pierre-Yves Laffont, Zhile Ren, Xiaofeng Tao, Chao Qian, and James Hays.
\newblock Transient attributes for high-level understanding and editing of
  outdoor scenes.
\newblock {\em ACM Transactions on Graphics}, 33(4):149, 2014.

\bibitem{lee2015predicting}
Stefan Lee, Haipeng Zhang, and David~J Crandall.
\newblock Predicting geo-informative attributes in large-scale image
  collections using convolutional neural networks.
\newblock In {\em IEEE Winter Conference on Applications of Computer Vision},
  2015.

\bibitem{leung2010proximate}
Daniel Leung and Shawn Newsam.
\newblock Proximate sensing: Inferring what-is-where from georeferenced photo
  collections.
\newblock In {\em IEEE Conference on Computer Vision and Pattern Recognition},
  2010.

\bibitem{lin2013cross}
Tsung-Yi Lin, Serge Belongie, and James Hays.
\newblock Cross-view image geolocalization.
\newblock In {\em IEEE Conference on Computer Vision and Pattern Recognition},
  2013.

\bibitem{lin2015learning}
Tsung-Yi Lin, Yin Cui, Serge Belongie, and James Hays.
\newblock Learning deep representations for ground-to-aerial geolocalization.
\newblock In {\em IEEE Conference on Computer Vision and Pattern Recognition},
  2015.

\bibitem{luo2008event}
Jiebo Luo, Jie Yu, Dhiraj Joshi, and Wei Hao.
\newblock Event recognition: Viewing the world with a third eye.
\newblock In {\em ACM International Conference on Multimedia}, 2008.

\bibitem{matzen2014scene}
Kevin Matzen and Noah Snavely.
\newblock Scene chronology.
\newblock In {\em European Conference on Computer Vision}, 2014.

\bibitem{mihail2016sky}
Radu~P Mihail, Scott Workman, Zach Bessinger, and Nathan Jacobs.
\newblock Sky segmentation in the wild: An empirical study.
\newblock In {\em IEEE Winter Conference on Applications of Computer Vision},
  2016.

\bibitem{regmi2018cross}
Krishna Regmi and Ali Borji.
\newblock Cross-view image synthesis using conditional {GAN}s.
\newblock In {\em IEEE Conference on Computer Vision and Pattern Recognition},
  2018.

\bibitem{salem2019anything}
Tawfiq Salem, Connor Greenwell, Hunter Blanton, and Nathan Jacobs.
\newblock Learning to map nearly anything.
\newblock In {\em IEEE International Geoscience and Remote Sensing Symposium},
  2019.

\bibitem{salem2018soundscape}
Tawfiq Salem, Menghua Zhai, Scott Workman, and Nathan Jacobs.
\newblock A multimodal approach to mapping soundscapes.
\newblock In {\em IEEE International Geoscience and Remote Sensing Symposium},
  2018.

\bibitem{seresinhe2015quantifying}
Chanuki~Illushka Seresinhe, Tobias Preis, and Helen~Susannah Moat.
\newblock Quantifying the impact of scenic environments on health.
\newblock {\em Scientific reports}, 5:16899, 2015.

\bibitem{vgg}
Karen Simonyan and Andrew Zisserman.
\newblock Very deep convolutional networks for large-scale image recognition.
\newblock In {\em International Conference on Learning Representations}, 2015.

\bibitem{srivastava2019understanding}
Shivangi Srivastava, John~E Vargas-Mu{\~n}oz, and Devis Tuia.
\newblock Understanding urban landuse from the above and ground perspectives: A
  deep learning, multimodal solution.
\newblock {\em Remote Sensing of Environment}, 228:129--143, 2019.

\bibitem{tang2015improving}
Kevin Tang, Manohar Paluri, Li Fei-Fei, Rob Fergus, and Lubomir Bourdev.
\newblock Improving image classification with location context.
\newblock In {\em IEEE International Conference on Computer Vision}, 2015.

\bibitem{yfcc100m}
Bart Thomee, David~A Shamma, Gerald Friedland, Benjamin Elizalde, Karl Ni,
  Douglas Poland, Damian Borth, and Li-Jia Li.
\newblock {YFCC100M}: The new data in multimedia research.
\newblock {\em Communications of the ACM}, 59(2):64--73, 2016.

\bibitem{tian2017cross}
Yicong Tian, Chen Chen, and Mubarak Shah.
\newblock Cross-view image matching for geo-localization in urban environments.
\newblock In {\em IEEE Conference on Computer Vision and Pattern Recognition},
  2017.

\bibitem{wang2016walk}
Jing Wang, Yu Cheng, and Rogerio Schmidt~Feris.
\newblock Walk and learn: Facial attribute representation learning from
  egocentric video and contextual data.
\newblock In {\em IEEE International Conference on Computer Vision}, 2016.

\bibitem{wang2016tracking}
Jingya Wang, Mohammed Korayem, Saul Blanco, and David~J Crandall.
\newblock Tracking natural events through social media and computer vision.
\newblock In {\em ACM International Conference on Multimedia}, 2016.

\bibitem{wang2013observing}
Jingya Wang, Mohammed Korayem, and David Crandall.
\newblock Observing the natural world with flickr.
\newblock In {\em ICCV Workshop on Computer Vision for Converging
  Perspectives}, 2013.

\bibitem{weyand2016planet}
Tobias Weyand, Ilya Kostrikov, and James Philbin.
\newblock Planet-photo geolocation with convolutional neural networks.
\newblock In {\em European Conference on Computer Vision}, 2016.

\bibitem{workman2015geocnn}
Scott Workman and Nathan Jacobs.
\newblock On the location dependence of convolutional neural network features.
\newblock In {\em IEEE/ISPRS Workshop: EARTHVISION: Looking From Above: When
  Earth Observation Meets Vision}, 2015.

\bibitem{workman2020dynamic}
Scott Workman and Nathan Jacobs.
\newblock Dynamic traffic modeling from overhead imagery.
\newblock In {\em IEEE Conference on Computer Vision and Pattern Recognition},
  2020.

\bibitem{workman2015wide}
Scott Workman, Richard Souvenir, and Nathan Jacobs.
\newblock Wide-area image geolocalization with aerial reference imagery.
\newblock In {\em IEEE International Conference on Computer Vision}, 2015.

\bibitem{workman2017beauty}
Scott Workman, Richard Souvenir, and Nathan Jacobs.
\newblock Understanding and mapping natural beauty.
\newblock In {\em IEEE International Conference on Computer Vision}, 2017.

\bibitem{workman2017unified}
Scott Workman, Menghua Zhai, David Crandall, and Nathan Jacobs.
\newblock A unified model for near and remote sensing.
\newblock In {\em IEEE International Conference on Computer Vision}, 2017.

\bibitem{xie2011im2map}
Ling Xie and Shawn Newsam.
\newblock {IM2MAP}: Deriving maps from georeferenced community contributed
  photo collections.
\newblock In {\em ACM SIGMM International Workshop on Social Media}, 2011.

\bibitem{zhai2017crossview}
Menghua Zhai, Zachary Bessinger, Scott Workman, and Nathan Jacobs.
\newblock Predicting ground-level scene layout from aerial imagery.
\newblock In {\em IEEE Conference on Computer Vision and Pattern Recognition},
  2017.

\bibitem{zhai2018geotemporal}
Menghua Zhai, Tawfiq Salem, Connor Greenwell, Scott Workman, Robert Pless, and
  Nathan Jacobs.
\newblock Learning geo-temporal image features.
\newblock In {\em British Machine Vision Conference}, 2018.

\bibitem{zhou2017places}
Bolei Zhou, Agata Lapedriza, Aditya Khosla, Aude Oliva, and Antonio Torralba.
\newblock Places: A 10 million image database for scene recognition.
\newblock {\em IEEE Transactions on Pattern Analysis and Machine Intelligence},
  40(6):1452--1464, 2017.

\end{thebibliography}
}
\newpage
\null
\vskip .375in
\twocolumn[{%
\begin{center}
 \textbf{\Large Supplemental Material:\\Learning a Dynamic Map of Visual Appearance}
\end{center}
\vspace*{27pt}
}]

\setcounter{section}{0}
\setcounter{figure}{0}
\makeatletter
\renewcommand{\thefigure}{S\arabic{figure}}

\section{Dynamic Visual Attribute Maps}

\begin{figure*}
  \centering
  \setlength\tabcolsep{1pt}
  \begin{tabular}{l|ccc}
    & \emph{January} & \emph{April} & \emph{July} \\    
    \raisebox{.7\height}{\rotatebox{90}{\em sat+time}} &
    \includegraphics[width=.32\linewidth]{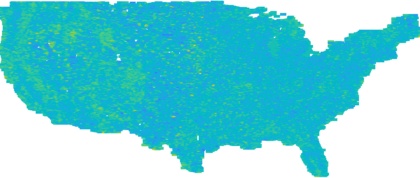} &
    \includegraphics[width=.32\linewidth]{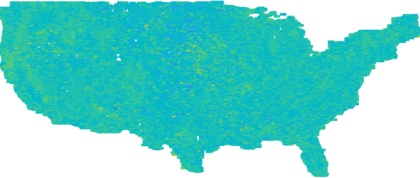} &
    \includegraphics[width=.32\linewidth]{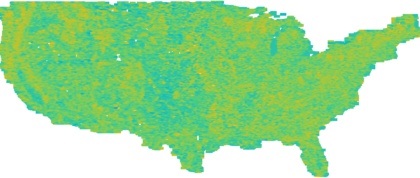}\\
    
    \raisebox{.2\height}{\rotatebox{90}{\em sat+time+loc}} &
    \includegraphics[width=.32\linewidth]{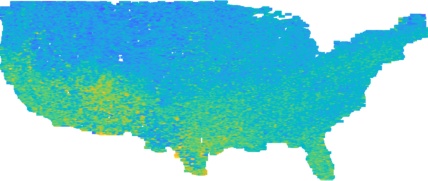} & 
    \includegraphics[width=.32\linewidth]{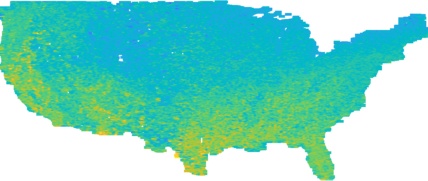} &
    \includegraphics[width=.32\linewidth]{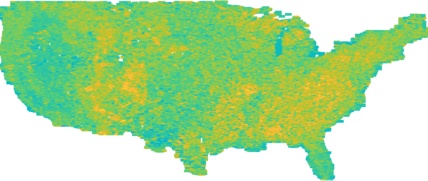}\\
    
  \end{tabular}
  \caption{Dynamic visual attribute maps over time for the transient attribute \emph{sunny}. In each, yellow (blue) corresponds to a higher (lower) value for the corresponding attribute.}
  \label{fig:maps-sunny}
\end{figure*}

\begin{figure*}
  \centering
  \setlength\tabcolsep{1pt}
  \begin{tabular}{l|ccc}
    & \emph{January} & \emph{April} & \emph{July} \\    
    \raisebox{.7\height}{\rotatebox{90}{\em sat+time}} &
    \includegraphics[width=.32\linewidth]{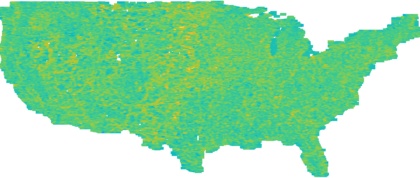} &
    \includegraphics[width=.32\linewidth]{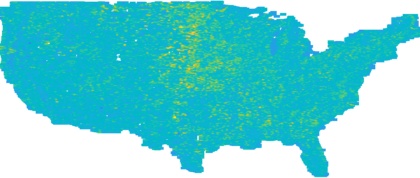} &
    \includegraphics[width=.32\linewidth]{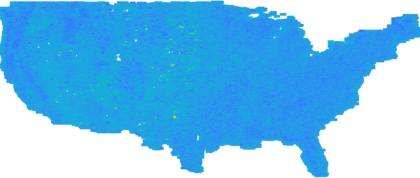}\\
    
    \raisebox{.2\height}{\rotatebox{90}{\em sat+time+loc}} &
    \includegraphics[width=.32\linewidth]{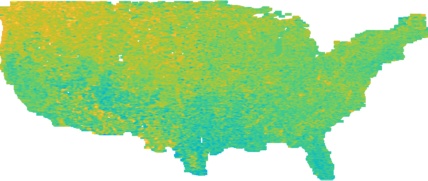} & 
    \includegraphics[width=.32\linewidth]{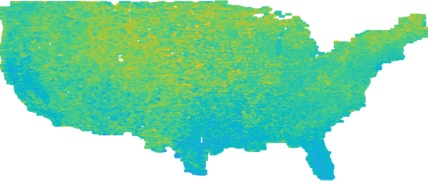} &
    \includegraphics[width=.32\linewidth]{supplementary/map_trans_diff_methods/stressful/combine/3_16.jpg}\\
    
  \end{tabular}
  \caption{Dynamic visual attribute maps over time for the transient attribute \emph{stressful}. In each, yellow (blue) corresponds to a higher (lower) value for the corresponding attribute.}
  \label{fig:maps-stressful}
\end{figure*}

We show additional dynamic attribute maps rendered from our model. See \figref{maps-sunny} for examples of the  \emph{sunny} attribute and \figref{maps-stressful} for examples of the \emph{stressful} attribute. For both attributes, we show our approach (\emph{sat+time+loc}) and a baseline that does not incorporate location as an input (\emph{sat+time}). For each, we specified the time of day as 4pm, and show three different months. In both models, we observe trends that match our expectations. For example, there tends to be more sunshine at 4pm in July than in January. However, the \emph{sat+time+loc} model does a better job of capturing large-scale spatial trends, such as the difference between the \emph{sunny} attribute in the north and south during January and April.

\section{Application: Image Localization}

We  evaluated the accuracy of our approach for the task of image geolocalization (Table 2 in the main paper). To summarize our method, we extracted the visual attributes of a query image and compared them against the visual attributes of an overhead image reference database, computed using the timestamp of the query image. To support this experiment, we created a new evaluation dataset that includes timestamps.
The results show that our model, \emph{sat+time+loc}, performs the best using all scoring strategies. 

\begin{figure*}[t!]
  \centering
  \setlength\tabcolsep{1pt}
  \begin{tabular}{l|ccc}
    & Feb. 3:00pm (UTC) & Nov. 4:00pm (UTC) &  Feb. 3:40pm(UTC) \\  
    
    \raisebox{1\height}{\rotatebox{90}{\em Query}} &
    \includegraphics[height=.15\linewidth]{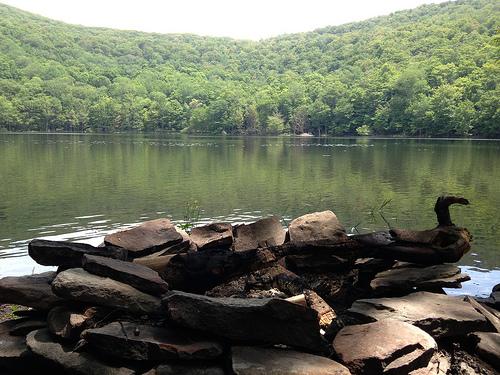}&
    \includegraphics[height=.15\linewidth]{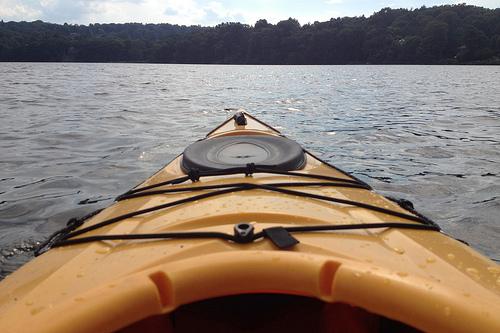}&
    \includegraphics[height=.15\linewidth]{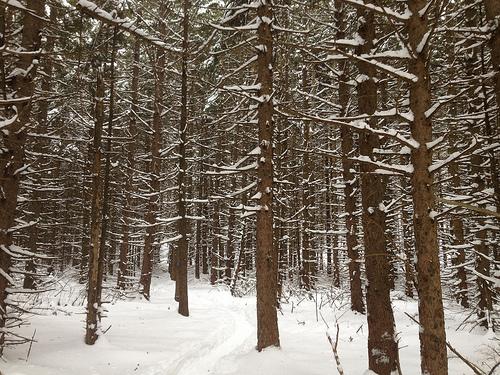} \\
    
    \raisebox{.6\height}{\rotatebox{90}{\em Transient}} &
    \includegraphics[width=.32\linewidth]{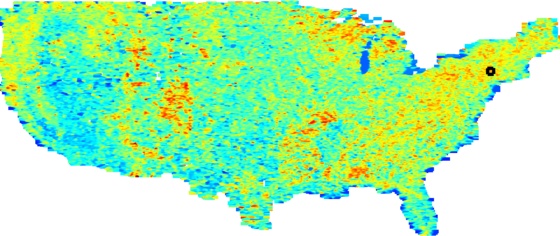} &
    \includegraphics[width=.32\linewidth]{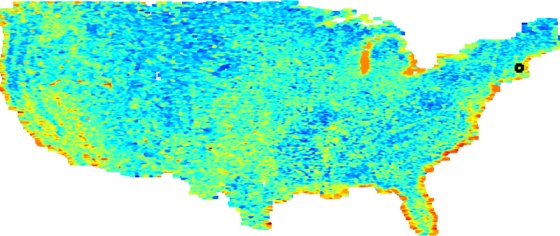} &
    \includegraphics[width=.32\linewidth]{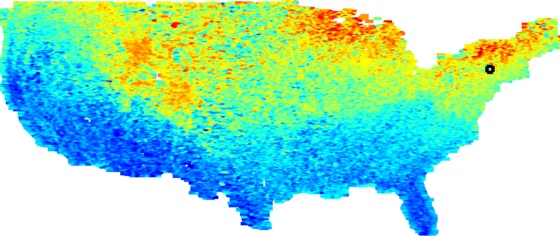} \\
    
    \raisebox{1\height}{\rotatebox{90}{\em Places}} &
    \includegraphics[width=.32\linewidth]{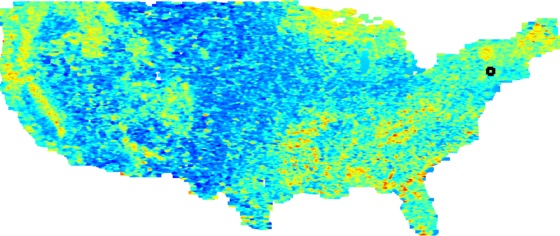} &
    \includegraphics[width=.32\linewidth]{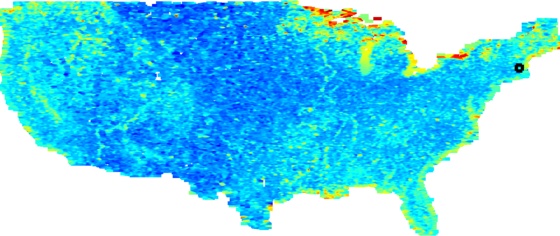} &
    \includegraphics[width=.32\linewidth]{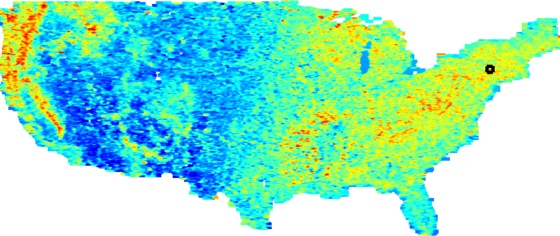} \\
    
    \raisebox{.5\height}{\rotatebox{90}{\em Combine}} &
    \includegraphics[width=.32\linewidth]{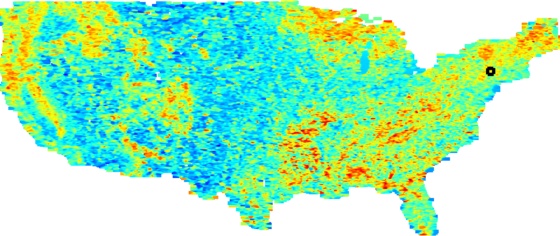} &
    \includegraphics[width=.32\linewidth]{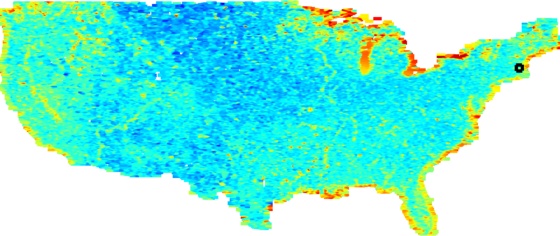} &
    \includegraphics[width=.32\linewidth]{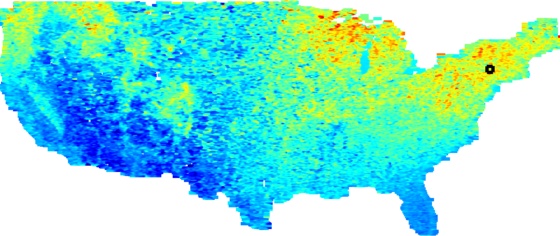} \\
    
    \raisebox{1\height}{\rotatebox{90}{\em Query}} &
    \includegraphics[height=.15\linewidth]{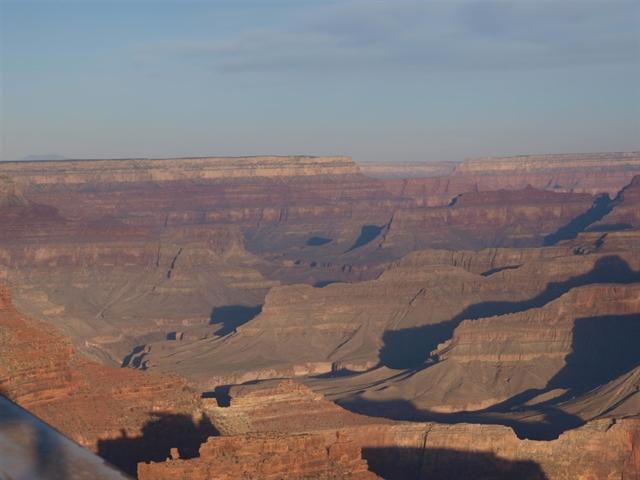}&
    \includegraphics[height=.15\linewidth]{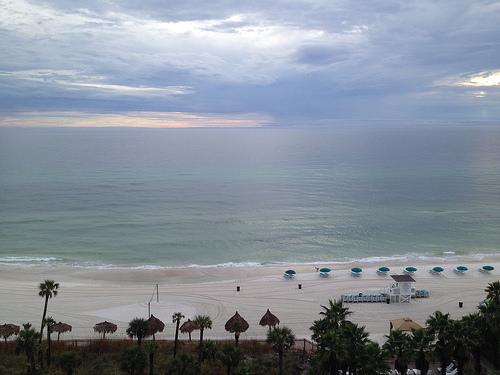} &
    \includegraphics[height=.15\linewidth]{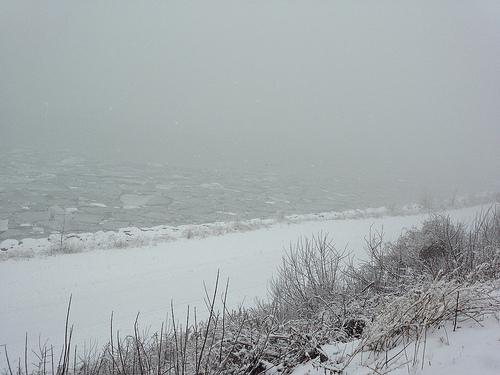} \\
    
    \raisebox{.6\height}{\rotatebox{90}{\em Transient}}&
    \includegraphics[width=.32\linewidth]{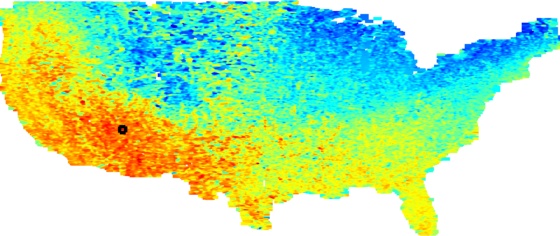} &
    \includegraphics[width=.32\linewidth]{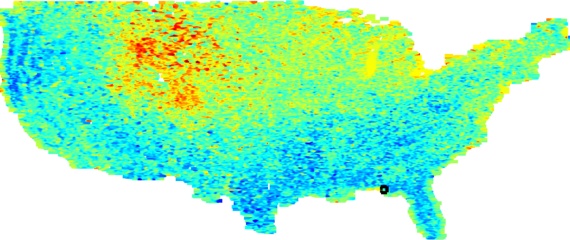} &
    \includegraphics[width=.32\linewidth]{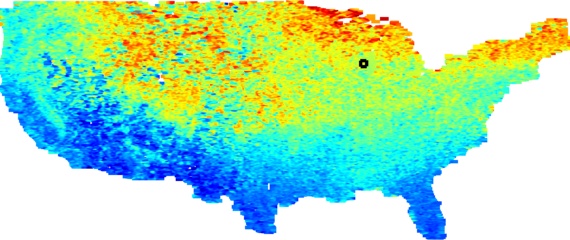} \\
    
    \raisebox{1\height}{\rotatebox{90}{\em Places}} &
    \includegraphics[width=.32\linewidth]{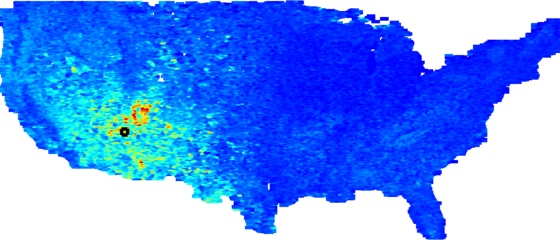} &
    \includegraphics[width=.32\linewidth]{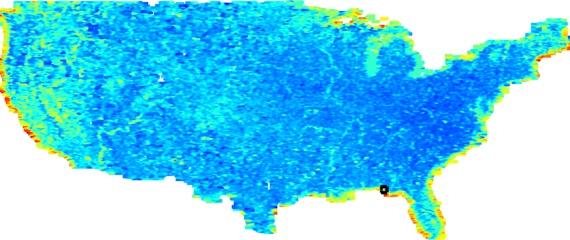} &
    \includegraphics[width=.32\linewidth]{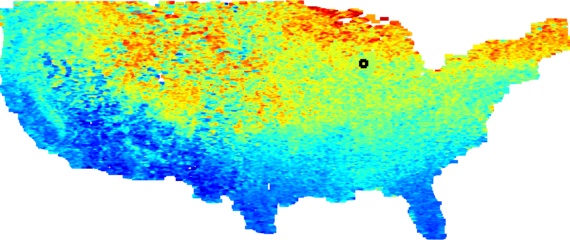} \\
    
    \raisebox{.5\height}{\rotatebox{90}{\em Combine}} &
    \includegraphics[width=.32\linewidth]{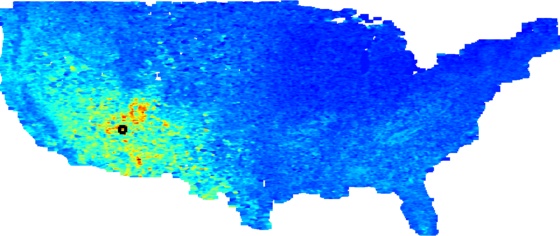} &
    \includegraphics[width=.32\linewidth]{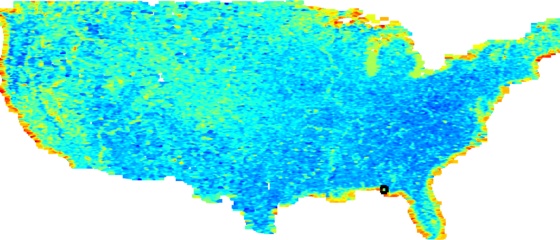} &
    \includegraphics[width=.32\linewidth]{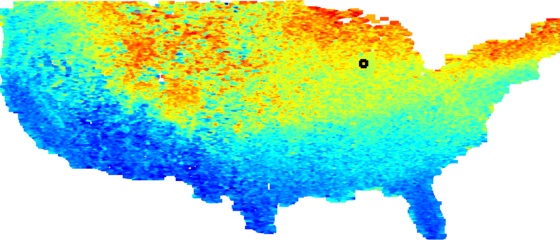} \\ 
    
  \end{tabular}
  \caption{Given a query ground-level image (top), we show localization results (bottom) for different scoring strategies, visualized as a heatmap. Red (blue) represents a higher (lower) likelihood that the image was captured at that location. }
  \label{fig:localization} 
\end{figure*}

In \figref{localization} we show qualitative localization results generated by our approach. For this experiment, we used \num{488224} overhead images from CVUSA as our reference database. The heatmap represents the likelihood that an image was captured at a specific location, where red (blue) is more (less) likely. Additionally, we compare the different scoring strategies on each row. Similar to our quantitative results, using the \emph{Combine} score produces heatmaps that more closely match the true location of the ground-level image.

\section{Application: Metadata Verification}

\begin{figure*}[!b]
  \centering
  \setlength\tabcolsep{1pt}
  \begin{tabular}{l|cc}
    
    &
    \includegraphics[height=3.4cm]{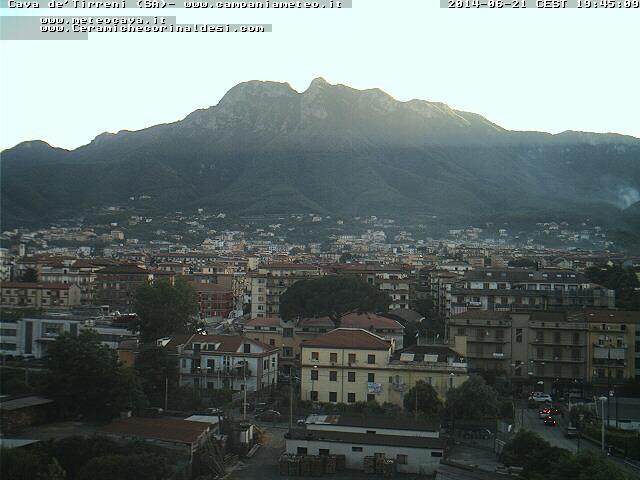}
    \includegraphics[height=3.4cm]{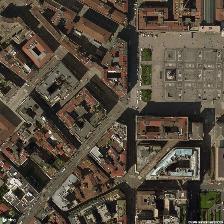} & 
    \includegraphics[height=3.4cm]{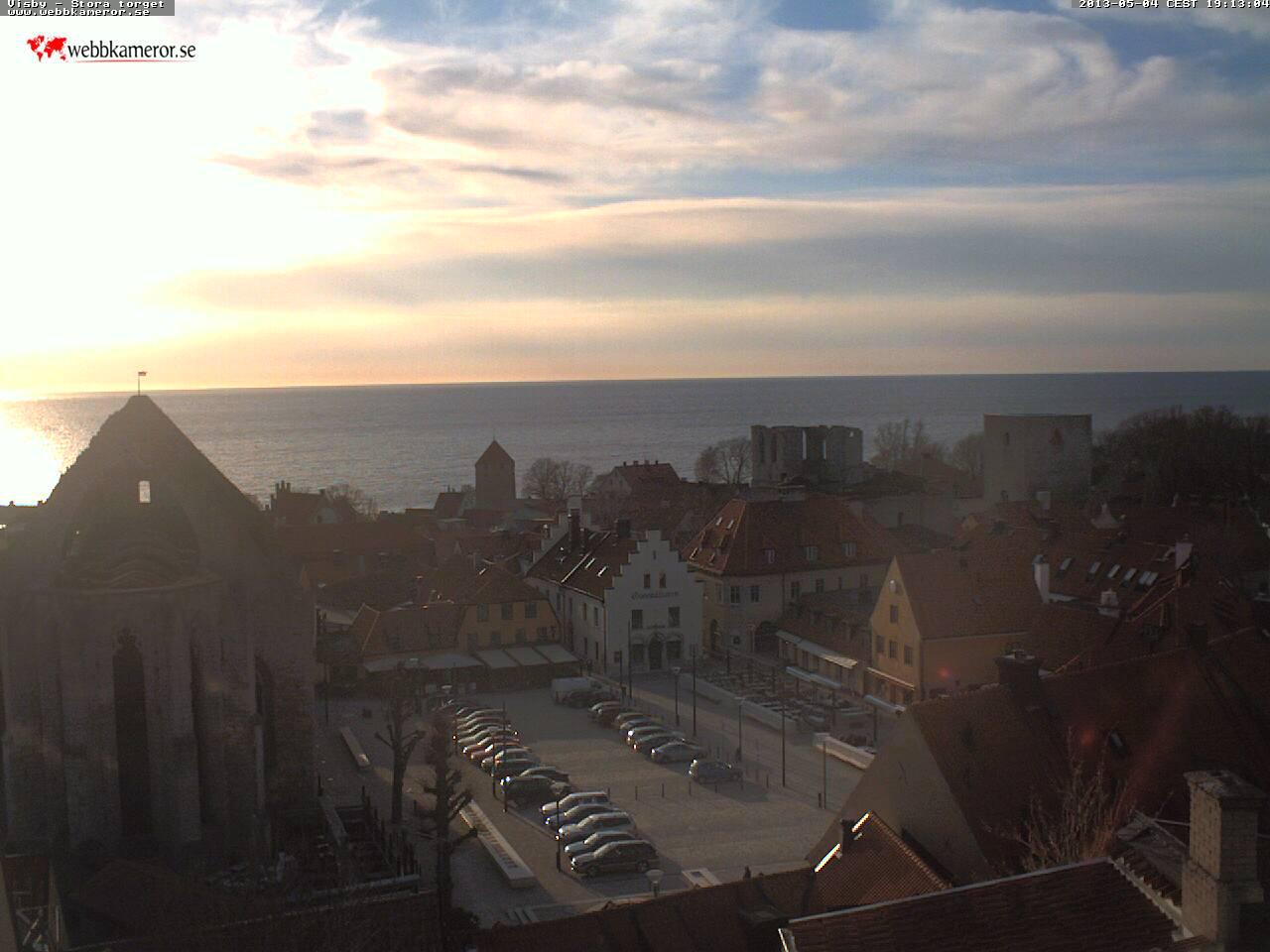}
    \includegraphics[height=3.4cm]{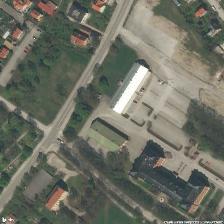} \\
    
    \raisebox{1.4\height}{\rotatebox{90}{\em time+loc}} &
    \includegraphics[width=.48\linewidth]{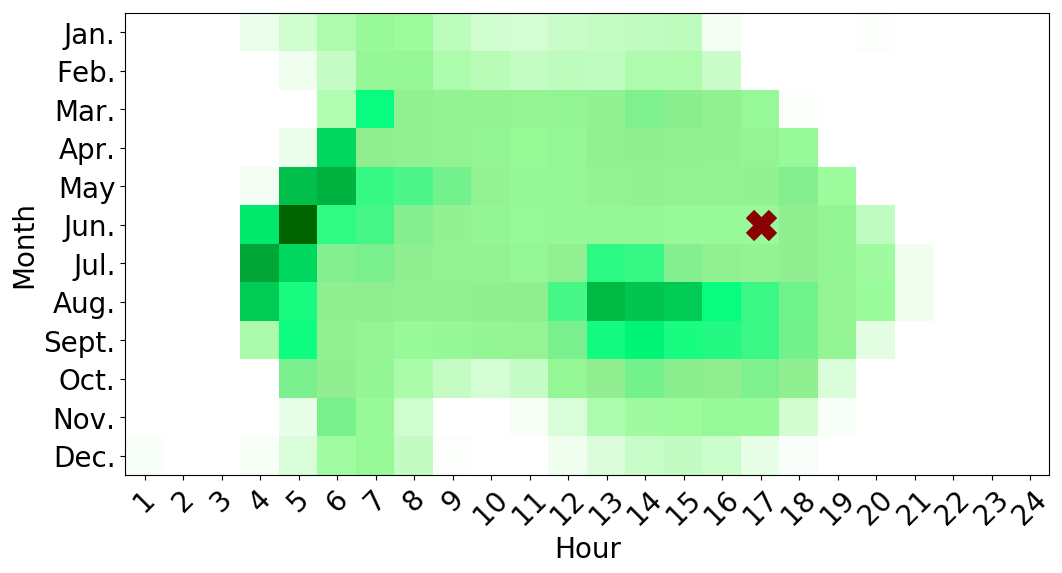} &
    \includegraphics[width=.48\linewidth]{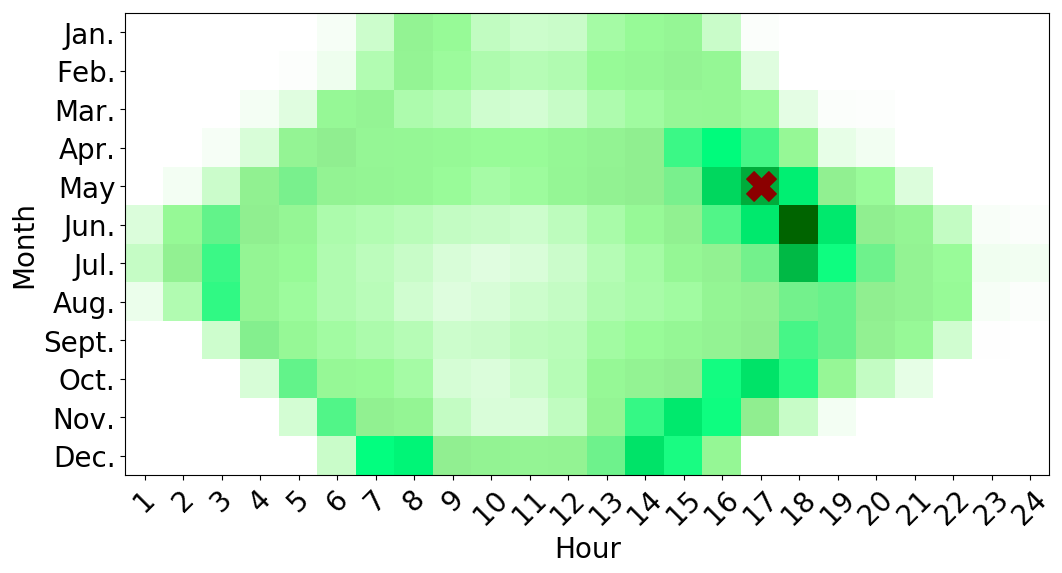} \\
    
    \raisebox{1.6\height}{\rotatebox{90}{\em sat+time}} &
    \includegraphics[width=.48\linewidth]{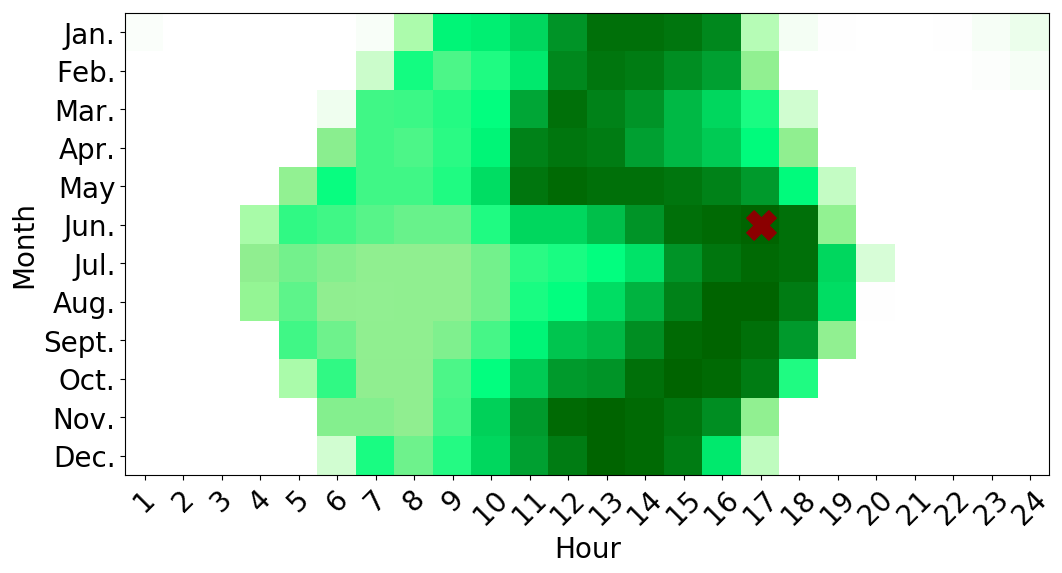} &
    \includegraphics[width=.48\linewidth]{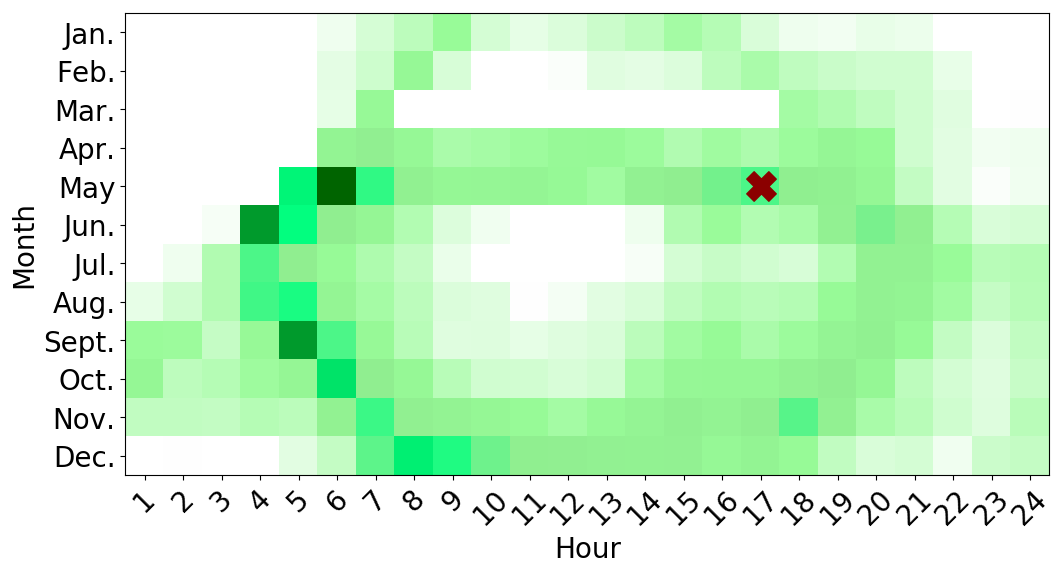} \\
    
    \raisebox{.8\height}{\rotatebox{90}{\em sat+time+loc}} &
    \includegraphics[width=.48\linewidth]{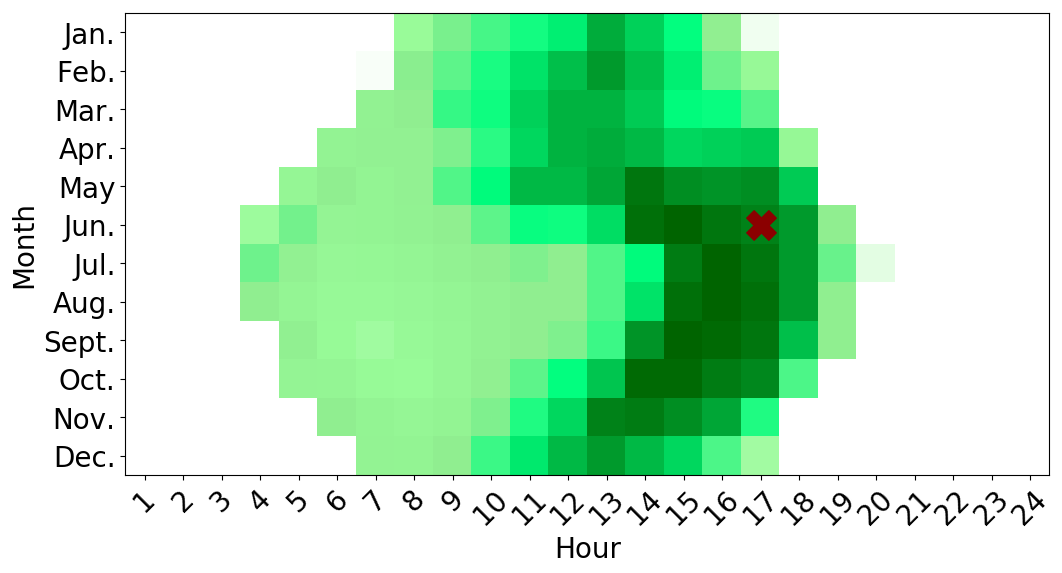} & 
    \includegraphics[width=.48\linewidth]{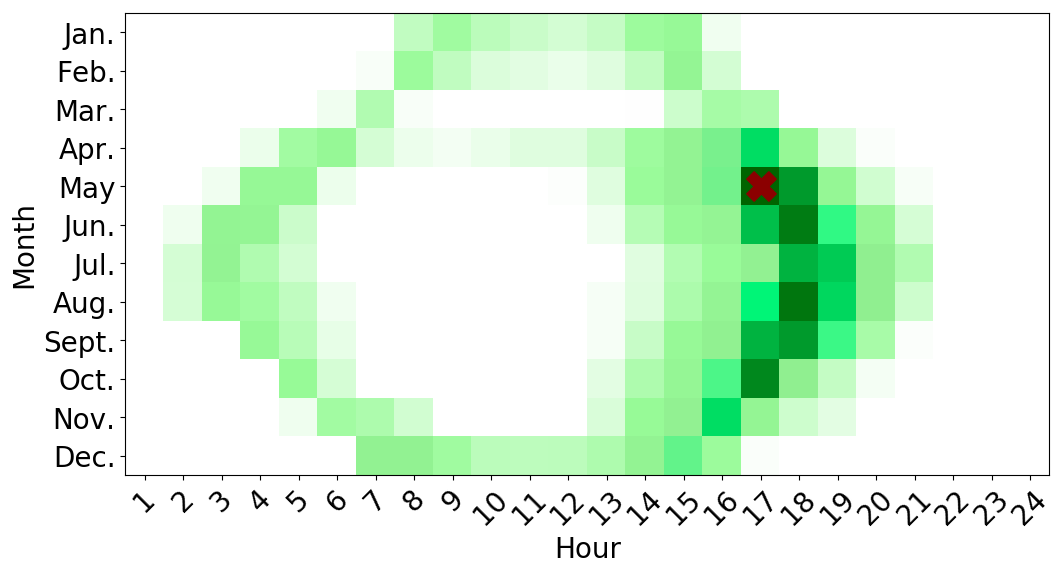} \\
  \end{tabular}
   
     \caption{Different examples highlighting temporal patterns learned by our model. (top) For each example, we show the original image and the overhead image of its location. (bottom) For every possible hour and month, we use 
  the different models (left) to predict the visual attributes. The heatmaps show the distance between the true and predicted visual attributes, with dark green (white) representing smaller (larger) distances.
  }

  \label{fig:supp_forensics_1}
   
 \end{figure*}

\begin{figure*}[htb]
  \centering
  \setlength\tabcolsep{1pt}
  \begin{tabular}{l|cc}
    &
    \includegraphics[height=3.4cm]{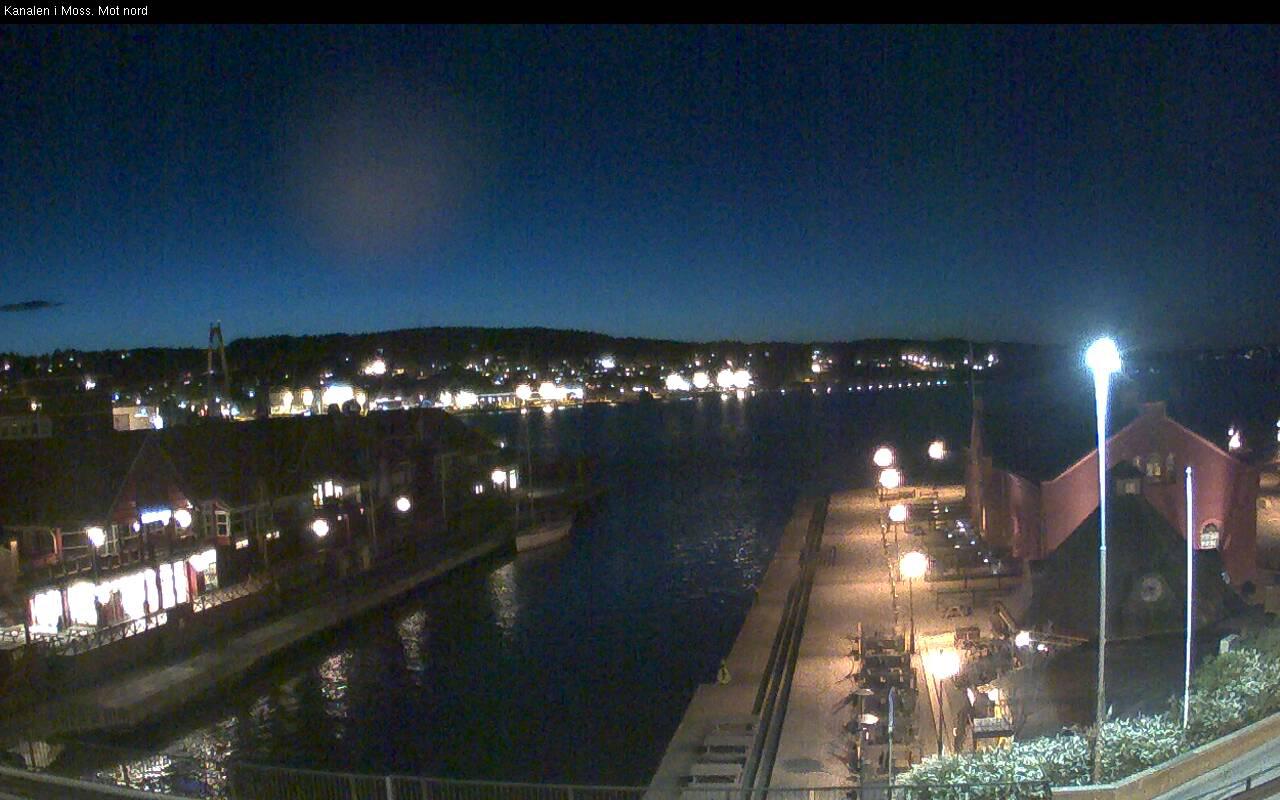}
    \includegraphics[height=3.4cm]{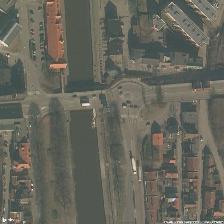} & 
    \includegraphics[height=3.4cm]{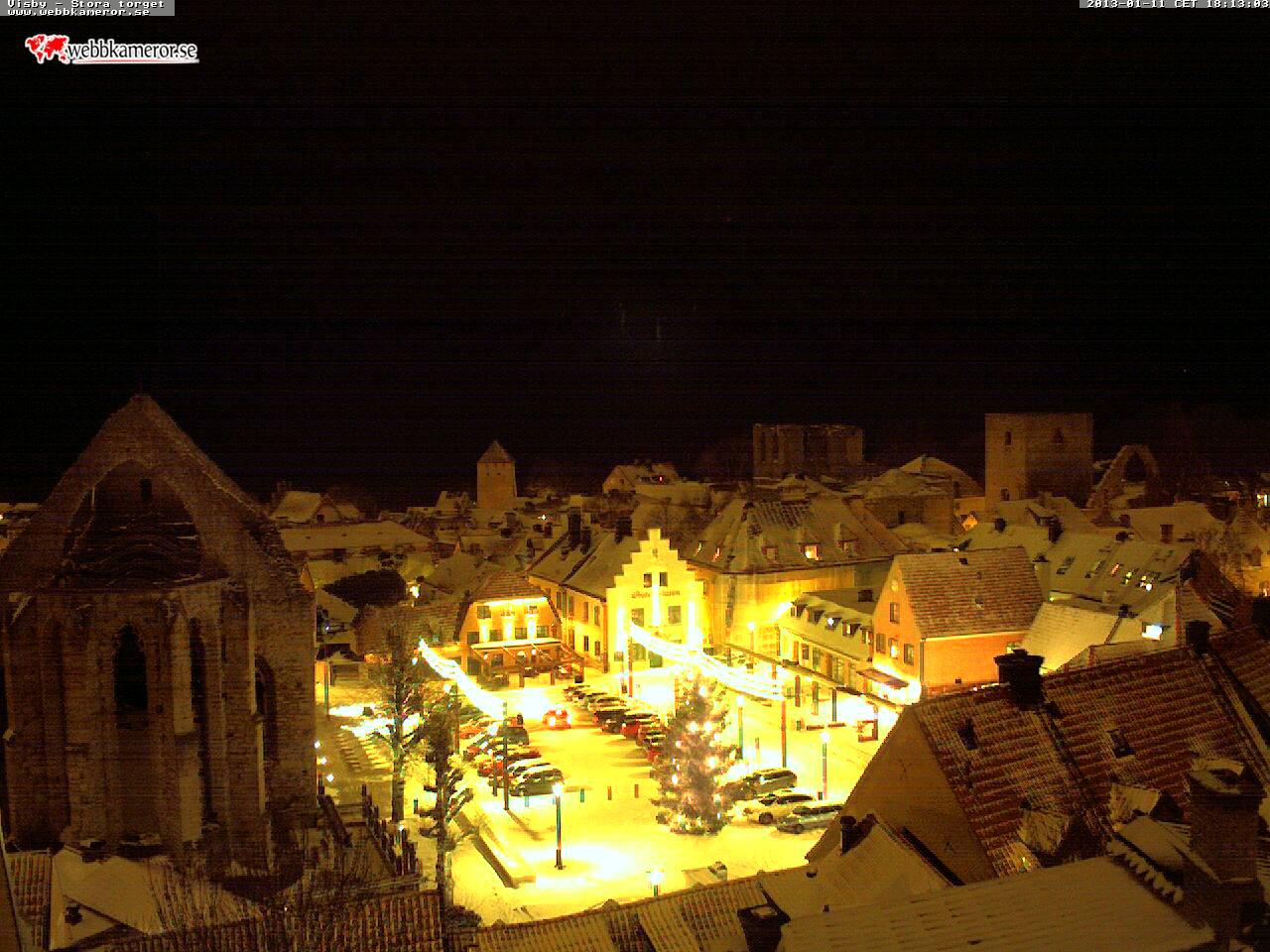}
    \includegraphics[height=3.4cm]{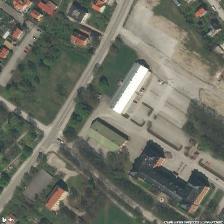} \\
    
    \raisebox{1.4\height}{\rotatebox{90}{\em time+loc}} &
    \includegraphics[width=.48\linewidth]{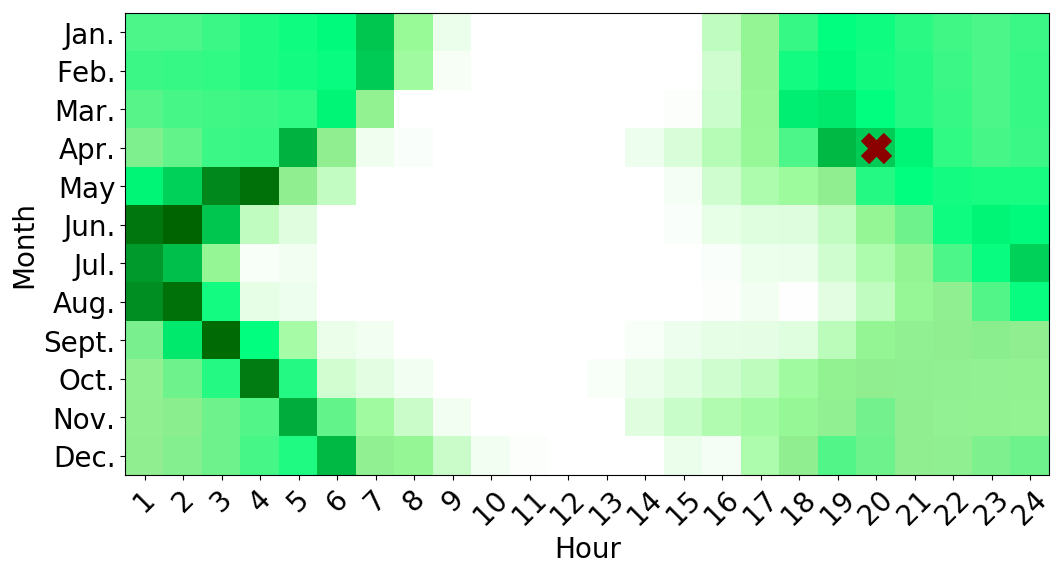} &
    \includegraphics[width=.48\linewidth]{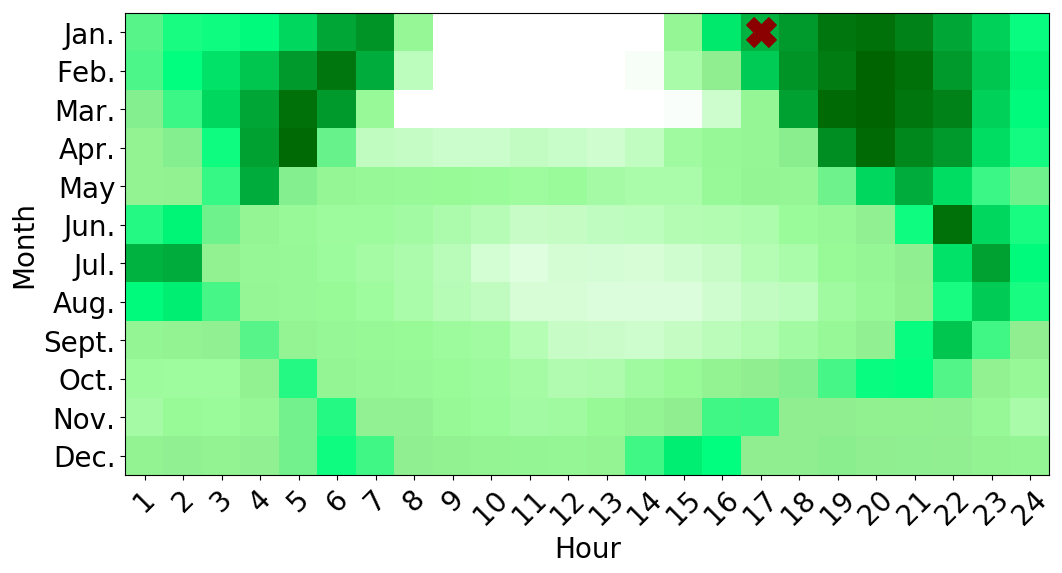} \\
    
    \raisebox{1.6\height}{\rotatebox{90}{\em sat+time}} &
    \includegraphics[width=.48\linewidth]{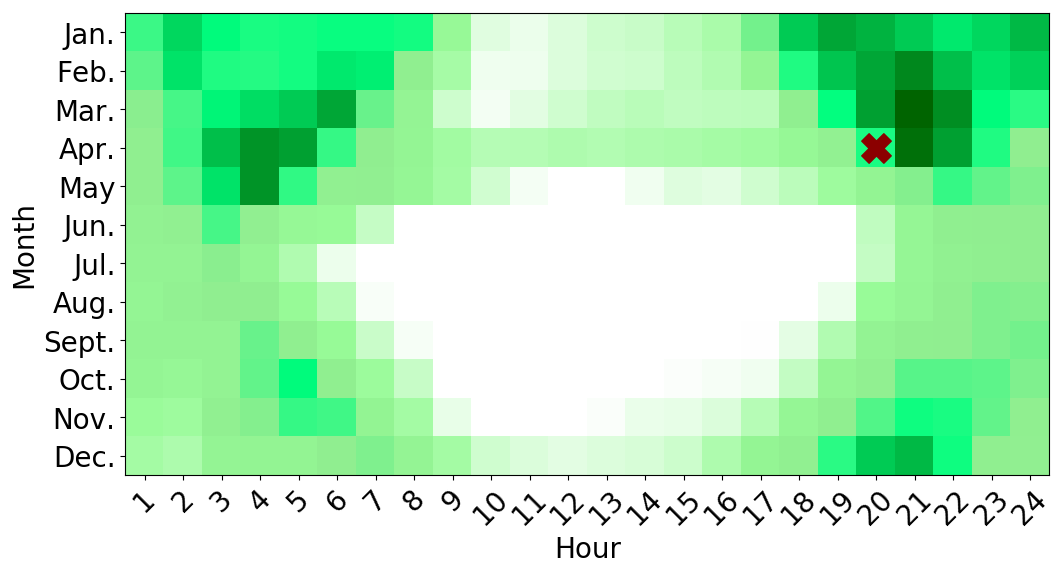} &
    \includegraphics[width=.48\linewidth]{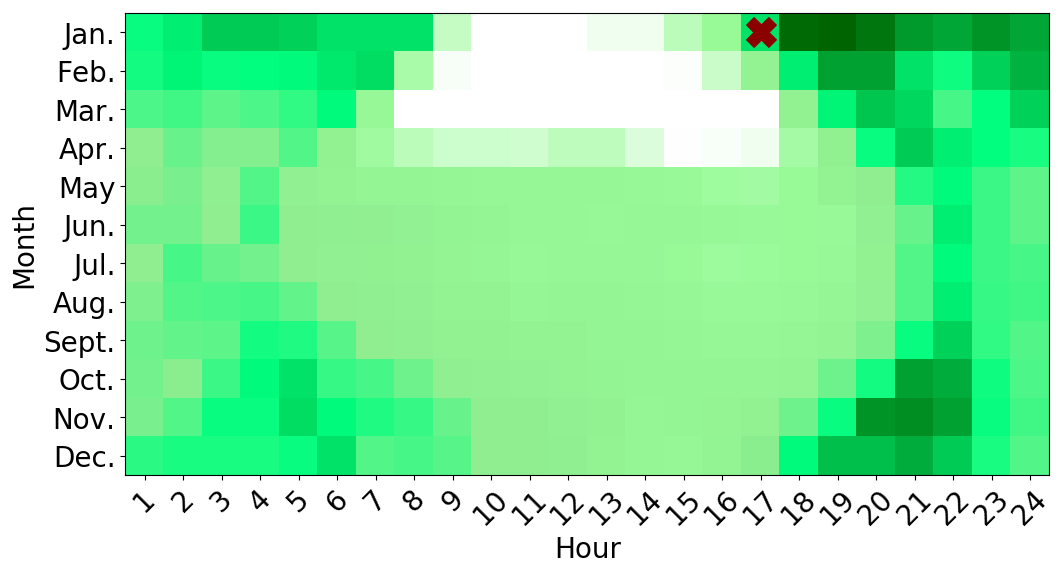} \\
    
    \raisebox{.8\height}{\rotatebox{90}{\em sat+time+loc}} &
    \includegraphics[width=.48\linewidth]{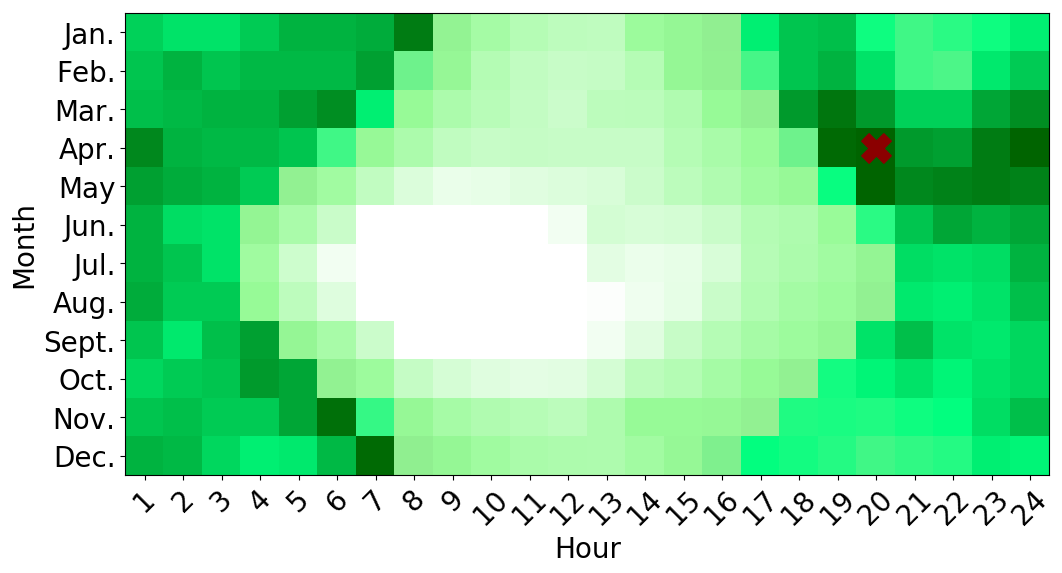} & 
    \includegraphics[width=.48\linewidth]{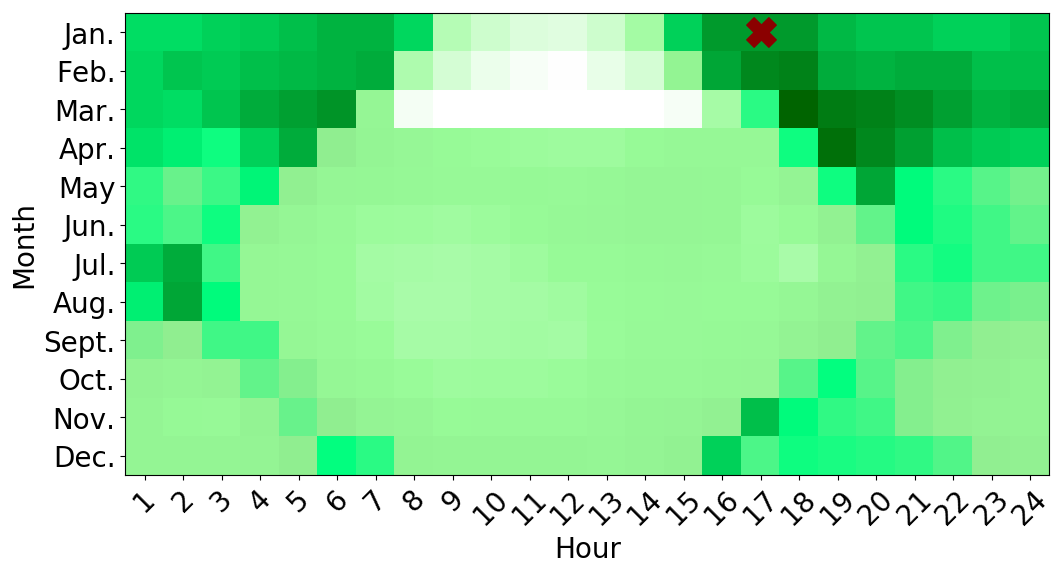} \\

  \end{tabular}

  \caption{Different examples highlighting temporal patterns learned by our model. (top) For each example, we show the original image and the overhead image of its location. (bottom) For every possible hour and month, we use 
  the different models (left) to predict the visual attributes. The heatmaps show the distance between the true and predicted visual attributes, with dark green (white) representing smaller (larger) distances.
  }

  \label{fig:supp_forensics_2}

\end{figure*}

For time verification accuracy, Table 3 in the main paper demonstrates that our approach, {\em sat+time+loc}, outperforms all baseline methods. In \figref{supp_forensics_1} and \figref{supp_forensics_2}, we show additional qualitative results for this task. The heatmaps reflect the distance between the visual attribute extracted from the ground-level image and the predicted attributes from the overhead image (varying the input time). This results in a distance for each possible time. The true capture time is indicated by the red $X$. As observed, our approach more accurately estimates the capture time of the ground-level image. 

\section{Discussion}

Our model combines overhead imagery, time, and geographic location to predict visual attributes. We have demonstrated the superiority of this combination, but we think there are several questions that naturally arise when considering our model. Here we provide answers, which we believe are supported by the evaluation.

\paragraph{Why do we need overhead imagery when it just depends on the location?} If our model was only dependent on geographic location, then we would need to learn a mapping between geographic location and the visual attribute. Consider something as simple as, ``does this geographic location contain a road?''. This would be a very complicated function to approximate using a neural network and we have seen that it does not work well. In contrast, it is relatively easy to estimate this type of information from the overhead imagery.

\paragraph{Why do we need to include geographic location if we have overhead imagery?} We think it makes it easier to learn larger scale trends, especially those that relate to time.  For example, the relationship between day length and latitude.  If we didn't include latitude we would have to estimate it from the overhead imagery, which would likely be highly uncertain.

\paragraph{Why don't we need an overhead image for each time?} The overhead image provides information about the type of place. This is unlike a satellite weather map, which would tell us what the conditions are at a particular time. While we do lose some information, this is accounted for by including geographic location and time as additional context. In practice it is best if the overhead image is captured relatively close in time (within a few years) to account for major land use and land cover changes.

\paragraph{Limitations} One of the limitations of this study is the reliance on social media imagery. This means that our visual appearance maps will exhibit biases about when people prefer to take pictures, or are willing to share pictures. For example, we are likely undersampling cold and stormy weather conditions and oversampling sunsets. This is part of the motivation for incorporating imagery from the AMOS dataset. This, at least, doesn't have the same temporal bias because the webcams collect images on a regular interval, regardless of conditions.  However, these are sparsely distributed spatially and, at least in our dataset, outnumbered by the social media imagery. Despite this, we were still able to demonstrate effective learning and this problem could be overcome as more data becomes available. Another limitation is that our current approach cannot model longer-term, year-over-year trends in visual attributes. This results because our representation of time only reflects the month and time of day, not the year. 

\end{document}